\documentclass[twocolumn]{autart}

\usepackage{amssymb,amsmath}
\usepackage{algorithm}
\usepackage{color}
\usepackage{hyperref}
\usepackage{graphicx}
\usepackage{tabularx}

\graphicspath{{figures/}}

\newcommand{\beq}{\begin{equation}}
\newcommand{\eeq}{\end{equation}}
\newcommand{\beqar}{\begin{eqnarray}}
\newcommand{\eeqar}{\end{eqnarray}}
\newcommand{\beqarno}{\begin{eqnarray*}}
\newcommand{\eeqarno}{\end{eqnarray*}}
\newcommand{\smallmat}[1]{\left[ \begin{smallmatrix}#1 \end{smallmatrix} \right]}

\newcolumntype{C}[1]{>{\centering\arraybackslash}p{#1}}
\newcommand{\rr}{{\mathbb R}}
\newcommand{\LL}{\mathcal{L}}
\newcommand{\TT}{\mathcal{T}}
\newcommand{\XX}{\mathcal{X}}
\newcommand{\YY}{\mathcal{Y}}
\newcommand{\KK}{\mathcal{K}}
\newcommand{\NN}{\mathcal{N}}

\newcommand{\rrinf}{{\mathbb R}\cup\{+\infty\}}

\newcommand{\LLinit}{\LL^{\rm init}}
\newcommand{\LLmode}{\LL^{\rm mode}}
\newcommand{\LLtrans}{\LL^{\rm trans}}

\newcommand{\matrice}[2]{\left[\hspace*{-.1cm}\begin{array}{#1} #2 \end{array}\hspace*{-.1cm}\right]}

\newcommand{\card}{\mathop{\#}\nolimits}
\newcommand{\sign}{\mathop{\rm sign}\nolimits}
\newcommand{\argmin}{\mathop{\rm arg\ min}\nolimits}
\newcommand{\argmax}{\mathop{\rm arg\ max}\nolimits}
\newcommand{\st}{\mbox{s.t.}}
\newcommand{\hyperPWA}{{\rho}}

\usepackage{enumitem}
\renewcommand{\theenumi}{\arabic{enumi}}

\renewcommand{\theenumii}{\arabic{enumii}}

\renewcommand{\theenumiii}{\arabic{enumiii}}

\definecolor{red}{rgb}{0.7,0,0}

\definecolor{green}{rgb}{0,0.4,0}

\definecolor{blue}{rgb}{0,0,0.7}

\newtheorem{proposition}{Proposition}
\newtheorem{corollary}{Corollary}

\begin{document}

\begin{frontmatter}
\runtitle{Fitting Jump Models}

\title{Fitting Jump Models}

\author[IMT]{A. Bemporad\thanksref{footnoteinfo}},
\author[POLIMI]{V. Breschi},
\author[IDSIA]{D. Piga},
\author[STANFORD]{S. Boyd}

\address[IMT]{IMT School for Advanced Studies Lucca, Piazza San Francesco 19, 55100 Lucca, Italy. Email: \texttt{alberto.bemporad@imtlucca.it}}
\address[POLIMI]{Dipartimento di Elettronica, Informazione e Bioingegneria, Politecnico di Milano, Piazza L.Da Vinci, 32, 20133 Milano, Italy. Email: \texttt{valentina.breschi@polimi.it}}
\address[IDSIA]{Dalle Molle Institute for Artificial Intelligence Research - USI/SUPSI, Galleria 2, Via Cantonale 2c, CH-6928 Manno, Switzerland. Email: \texttt{dario.piga@supsi.ch}}
\address[STANFORD]{Department of Electrical Engineering, Stanford University, Stanford CA 94305, USA. Email: \texttt{boyd@stanford.edu}}

\thanks[footnoteinfo]{Corresponding author.}


\begin{abstract}
We describe a new framework for fitting jump models to a sequence of data. The key idea is to alternate between minimizing a loss function to fit multiple model parameters, and minimizing a discrete loss function to determine which set of model parameters is active at each data point. The framework is quite general and encompasses popular classes of models, such as hidden Markov models and piecewise affine models. The shape of the chosen loss functions to minimize
determine the shape of the resulting jump model.
\end{abstract}

\end{frontmatter}

\section{Introduction}
In many regression and classification problems the training dataset is formed by input and output observations with time stamps. However, when fitting the function that maps input data to output data, most algorithms used in supervised learning do not take the temporal order of the data into account. For example, in linear regression problems solved by least squares $\min_\theta \|A\theta-b\|_2^2$ each row of $A$ and $b$ is associated with a data-point, but clearly the solution $\theta^\star$ is the same no matter how the rows of $A$ and $b$ are ordered. In system identification temporal information is often used only to construct the input samples (or regressors) and outputs, but then it is neglected. For example, in estimating autoregressive models with exogenous inputs (ARX), the regressor is a finite collection of current and past signal observations, but the order of the regressor/output pairs is irrelevant when least squares are used. Similarly, in logistic regression and support vector machines the order of the data points does not affect the result. In training forward neural networks using stochastic gradient descent, the samples may be picked up randomly (and more than once) by the solution algorithm, and again their original temporal ordering is neglected. 

On the other hand, there are many applications in which relevant information is contained not only in data values but also in their temporal order. In particular, if the time each data-point was collected is taken into account, one can detect changes in the type of regime the data were produced. 
Examples range from video segmentation~\cite{OhSa2008,chan2008modeling} to speech recognition~\cite{Rab89,ShWo2008}, asset-price models in finance~\cite{TiMar15,Gu2011markov},  human action classification~\cite{PaReMac2001,OzSzLa2010}, and many others. All these examples are characterized by the need of fitting multiple models and understanding when switches from one model to another occur.

Piecewise affine (PWA) models attempt at fitting multiple affine models to a dataset, where each model is active based on the location of the input sample in a polyhedral partition of the input space~\cite{FeMuLiMo03,BPB16a}. However, as for ARX models, the order of the data is not relevant in computing the model parameters and the polyhedral partition. In some cases, mode transitions are captured by finite state machines, for example in hybrid dynamical models with logical states, where the current mode and the next logical state are generated deterministically by Boolean functions~\cite{BG06,BrBePi16ECC}. In spite of the difficulty of assessing whether a switched linear dynamical system is identifiable from input/output data~\cite{VCS02}, a rich variety of identification methods have been proposed in the literature~\cite{FeMuLiMo03,BRL01a,JWH04,BGPV05,JHFVPN05,BPB16a,Pil16}.

Hidden Markov models (HMMs) treat instead the mode as a stochastic discrete variable, whose temporal dynamics is described by a Markov chain~\cite{Rab89}. 
Natural extensions of hidden Markov models consider the cases in which each mode is associated with a linear function of the input~\cite{Fri94,costa2006discrete,OhLj13}. Hidden Markov models are usually trained using the Baum-Welch algorithm~\cite{BPSW70}, a forward-backward version of the more general Expectation Maximization (EM) algorithm~\cite{DLR77}.

In this paper we consider rather general \emph{jump models} to fit a temporal sequence of data that takes the ordering of the data into account. The proposed fitting algorithm alternates two steps: estimate the parameters of multiple models and estimate the temporal sequence of model activation, until convergence. 
The model fitting step can be carried out exactly when it reduces to a convex optimization problem, which is often the case. The mode-sequence step is always carried out optimally using dynamic programming. 

Our jump modeling framework is quite general. The structure of the model depends on the shape of the function that is minimized to obtain the model parameters, the way the model jumps depends on the function that is minimized to get the sequence of model activation. When we impose no constraints or penalty on the model sequence, our method reduces to automatically splitting the dataset in $K$ clusters and fitting
one model per cluster, which is 
a generalization of $K$-means \cite[Algorithm 14.1]{HTF09}. 
Hidden Markov models (HMMs) are a special case of jump models, as we will show in the paper. Indeed, jump models have broader descriptive capabilities than HMMs, for example the sequence of discrete states may not be necessarily generated by a Markov chain and could be a deterministic function. Moreover, as stated above, jump models can have rather arbitrary model shapes. 

After introducing jump models in Section~\ref{sec:models} and giving a statistical interpretation of the loss function in Section~\ref{sec:stoch-interpret}, we provide algorithms for fitting jump models
to data and to estimate output values and hidden modes from available input samples in Section~\ref{sec:algorithms}, emphasizing differences and analogies with HMMs. Finally, in Section~\ref{sec:examples}  we show four examples of application of our approach for regression and classification, using both synthetic and experimental data sets. 

The code implementing the algorithms described in the paper is available
at \url{http://cse.lab.imtlucca.it/~bemporad/jump_models/}.

\subsection{Setting and goal}
We are given a training sequence of data pairs $(x_t,y_t)$, $t=1,\ldots,T$, with $x_t\in\XX$, $y_t\in\YY$. We refer to $t$ as the \emph{time} or \emph{period}, $x_t$ as the \emph{regressor}
or \emph{input}, and $y_t$ as the \emph{outcome} or \emph{output} at time $t$. The training sequence is used to build a regression model that  provides a \emph{prediction} $\hat y_t$ of $y_t$
given the available inputs $x_1,\ldots,x_t$, and possibly past outputs $y_1,\ldots,y_{t-1}$.
We are specifically interested in models
where $\hat y_t$ is not simply a static function of $x_t$,
but rather we want to exploit the additional information embedded
in the temporal ordering of the data. As we will detail later,
our regression model is implicitly  defined by the minimization
of a \emph{fitting loss} $J$ that depends on $x_1,\ldots,x_t, y_1,\ldots,y_{t-1},y_t$ and other variables and parameters. The
chosen shape for $J$ determines the
structure of the corresponding regression model.

Given a production data sequence $(\tilde x_1,\tilde y_1),\ldots$, thought to be generated by a similar process that produced the training data, the quality of the regression model over a time period $t=1,\ldots,\tilde T$ will be judged by the average \emph{true loss}
\begin{equation}
L^{\rm true}=\frac{1}{\tilde  T}\sum_{t=1}^{\tilde T}\ell^{\rm true}(\hat y_t,\tilde y_t)    
\label{eq:trueloss}
\end{equation}
where $\ell^{\rm true}:\YY\times \YY\to\rr$ penalizes the mismatch
between $\hat y_t$ and $\tilde y_t$, with $\ell(y,y)=0$ for all $y\in\YY$.

\section{Regression models}
\label{sec:models}
\subsection{Single model}
A simple form of deriving a regression model is to introduce a \emph{model parameter} $\theta\in\rr^{d}$,
a \emph{loss function} $\ell:\XX\times\YY\times \rr^d\to\rrinf$, 
and a \emph{regularizer} $r:\rr^d\to\rrinf$ defining the \emph{fitting objective}  
\begin{subequations}
\begin{equation}
    J(X,Y,\theta)=\sum_{t=1}^T\ell(x_t,y_t,\theta)+r(\theta)
\label{eq:criterion0}
\end{equation}
where $X=(x_1,\ldots,x_T)$, $Y=(y_1,\ldots,y_T)$. 
For a given training data set $(X,Y)$, let
\begin{equation}
    \theta^\star=\argmin_\theta J(X,Y,\theta)    
\end{equation}
be the optimal model parameter. By fixing $\theta=\theta^\star$
and exploiting the separability of the loss $J$ in~\eqref{eq:criterion0}
we get the following regression model
\begin{eqnarray}
    \hat y_t&=&\argmin_{y} J(X,Y,\theta^\star)=
    \argmin_{y}\ell(x_t,y,\theta^\star)\nonumber\\
    &=:&\varphi(x_t)
\label{eq:predictor0}
\end{eqnarray}
\label{eq:model0}%
\end{subequations}
where $\varphi:\XX\to\YY$ as the regression model, with ties in the arg min broken arbitrarily. For example, when  $\ell(x_t,y,\theta)=\left\|y - \theta'x_t\right\|^2_2$ we get 
the standard linear regression model $\hat y_t = \theta'x_t$.

Model~\eqref{eq:model0} can be enriched by adding \emph{output information sets} $\YY_t\subseteq \YY$ that augment the information
that is available about $y_t$, 
\begin{equation}
    \hat y_t=\argmin_{y\in\YY_t}\ell(x,y,\theta^\star)
\label{eq:predictor0t}
\end{equation}
where $\YY_t=\YY$ if no extra information on $y_t$ is given. For example,
if we know a priori that $y_t\geq 0$ we can set $\YY_t$ equal to the nonnegative orthant.

\subsection{K-models}
Let us add more flexibility and introduce multiple model parameters $\theta_s\in\rr^{d}$, $s=1,\ldots,K$,
and a latent \emph{mode} variable $s_t$ that determines the model parameter $\theta_{s_t}$ that is active at step $t$. Fitting a \emph{K-model} on the training data set
$(X,Y)$, entails choosing the $K$ models by minimizing
\begin{equation}
    J(X,Y,\Theta,S)=\sum_{t=1}^T\ell(x_t,y_t,\theta_{s_t})+\sum_{i=1}^Kr(\theta_i)
\label{eq:criterion1}
\end{equation}
with respect to $\Theta=(\theta_1,\ldots,\theta_K)$ and $S=(s_1,\ldots,s_T)$. The optimal parameters $\theta_1^\star,\ldots,\theta_K^\star$ define the
$K$-model 
\begin{equation}
 (\hat y_t,\hat s_t)=\argmin_{y,s}\ell(x_t,y,\theta_{s}^\star).
\label{eq:K-models}
\end{equation}
Note that the objective function in~\eqref{eq:criterion1} is used to estimate the model parameters $\theta_1^\star,\ldots,\theta_K^\star$ based on the entire training dataset, while~\eqref{eq:K-models} defines the model used to infer the output $\hat y_t$
and discrete state $\hat s_t$ given the input $x_t$, as exemplified in the next section.

\subsubsection{K-means and piecewise affine models} 
The standard $K$-means model~\cite{HTF09} is obtained by setting $y_t=x_t$,
$r(\theta)=0$, and 
\begin{equation}
\ell(x_t,y_t,\theta_{s_t})=\frac{1}{2}\|y_t-\theta_{s_t}\|_2^2+\frac{1}{2}\|x_t-\theta_{s_t}\|_2^2
=\|x_t-\theta_{s_t}\|_2^2
\label{eq:kmeans}
\end{equation}
In this case, minimizing~\eqref{eq:criterion1} assigns each datapoint
$x_t$ to the cluster indexed by $s^\star_t$, and defines $\theta^\star_1,\ \ldots,\theta^\star_K$ 
as the centroids of the resulting $K$ clusters. Moreover, the regression model defined by~\eqref{eq:kmeans} returns 
\begin{equation}
    \hat s_t=\argmin_s\|\tilde x_t-\theta^\star_s\|_2^2,\quad \hat y_t=\theta_{\hat s_t}
\label{eq:styt-kmeans}
\end{equation}
that is the index $\hat s_t$ of the centroid $\theta^\star_{\hat s_t}$ which is closest to the given input $x_t$, and sets $\hat y_t=\theta^\star_{\hat s_t}$ as the best estimate of $x_t$. 

More generally, by setting
\begin{equation}
\ell(x_t,y_t,\theta_{s_t})=\|y_t-\theta_{y,s_t}'\smallmat{x_t\\1}\|_2^2+ \hyperPWA \|x_t-\theta_{x,s_t}\|_2^2
\label{eq:pwa-Voronoi}
\end{equation}
with $\theta_{s_t}=(\theta_{y,s_t}, \theta_{x,s_t})$ and $\rho>0$,  
we obtain a piecewise affine (PWA) model over the piecewise linear partition generated by the Voronoi diagram of $(\theta^\star_{x,1},\ldots,\theta^\star_{x,K})$, i.e., the regression model~\eqref{eq:K-models} becomes 
\begin{equation}  \label{eq:regr-ex1}
    \hat s_t=\argmin_s\|x_t-\theta^\star_{x,s}\|_2^2,\quad
    \hat y_t=(\theta^\star_{y,\hat s_t})'\smallmat{x_t\\1}
\end{equation}
The hyper-parameter $\hyperPWA$ in \eqref{eq:pwa-Voronoi} trades off between fitting the output $y_t$ and clustering the inputs $(x_1,\ldots,x_t)$ based on their mutual Euclidean distance.

A more general PWA model can be defined by  setting 
\begin{equation}
\begin{split}
\ell(x_t,y_t,\theta_{s_t})=&\|y_t-\theta_{y,s_t}'\smallmat{x_t\\1}\|_2^2\\&\hspace*{-1cm}+\hyperPWA
\sum_{\begin{smallmatrix}j=1\\j\neq s_t\end{smallmatrix}}^K\max\left\{0,(\theta_{x,j}-\theta_{x,s_t})'\smallmat{x_t\\1}+1\right\}^2
\end{split}
\label{eq:pwa}
\end{equation}
where $\max_s\{\theta_{x,s}'\smallmat{x\\1}\}$ defines a  piecewise linear separation function that induces a polyhedral partition of the input space~\cite{BM94,BPB16a}.
In this case it is immediate to verify that the regression model induced by~\eqref{eq:K-models}
is
\begin{equation} \label{eq:pwainference}
    \hat s_t=\argmax_s\{(\theta^\star_{x,s})'\smallmat{x_t\\1}\},\quad 
    \hat y_t=(\theta^\star_{y,\hat s_t})'\smallmat{x_t\\1}.
\end{equation}

\subsection{Jump model}
The models introduced above do not take into account the temporal order in which the samples $(x_t,y_t)$ are generated. To this end,
we add a \emph{mode sequence loss} $\LL$ in the fitting objective~\eqref{eq:criterion1}
 \begin{equation}
    J(X,Y,\Theta,S)=\sum_{t=1}^T\ell(x_t,y_t,\theta_{s_t})+\sum_{k=1}^Kr(\theta_k)+\LL(S),
\label{eq:criterion}
\end{equation}
where $S=(s_0,s_1,\ldots,s_T)$ is the mode sequence. We define
$\LL:\KK^{T+1}\to\rrinf$ in~\eqref{eq:criterion} as
\begin{subequations}
\begin{equation}
    \LL(S)=\LLinit(s_0)+\sum_{t=1}^T\LLmode(s_t)+ 
    \sum_{t=1}^T\LLtrans(s_t,s_{t-1})
    \label{eq:modeseqloss}
\end{equation}
where $\KK=\{1,\ldots,K\}$, $\LLinit:\KK\to\rrinf$
is the \emph{initial mode cost}, $\LLmode:\KK\to\rrinf$ is the \emph{mode cost},
and $\LLtrans:\KK^2\to\rrinf$
is the \emph{mode transition cost}. 
We discuss possible choices for $\LL$
in Sections~\ref{sec:mode-loss} and~\ref{sec:stoch-interpret}.

With a little abuse of notation, we write
\begin{equation}
    J(X,Y,\Theta,S)=\ell(X,Y,\Theta,S)+r(\Theta)+\LL(S)    
\label{eq:criterion2}
\end{equation}
where 
\begin{equation}
\ell(X,Y,\Theta,S)=\sum_{t=1}^T\ell(x_t,y_t,\theta_{s_t}),\quad
r(\Theta)=\sum_{k=1}^Kr(\theta_k).
\label{eq:ell-r}
\end{equation}
\label{eq:cost-form}%
\end{subequations}

As with any model, the choice of the fitting objective~\eqref{eq:cost-form}
should trade off between fitting the given data and prior assumptions we have about the models and the mode sequence. In particular, the mode sequence loss $\LL$ in~\eqref{eq:modeseqloss} takes into account the temporal 
structure of the mode sequence, for example that the mode might change (i.e., $s_t\neq s_{t-1}$) rarely.

A jump model can be used for several tasks
beyond inferring the values $\hat y_t$.
In \emph{anomaly identification}, we are interested in determining times $t$ for which the jump model
does not fit the data point $y_t$ well. In \emph{model change detection} we are interested
in identifying times $t$ for which $\hat s_t\neq \hat s_{t-1}$. In \emph{control systems} jump models can be used to approximate nonlinear/discontinuous dynamics and design model-based control policies, state estimators, and fault-detection algorithms.

\subsubsection{Mode loss functions}
\label{sec:mode-loss}
We discuss a few options for choosing the mode loss functions $\LLinit$, $\LLmode$, $\LLtrans$
defining the mode sequence loss $\LL$ in~\eqref{eq:modeseqloss}. As we assume that the number $K$ of possible modes must be fixed, $K$ must be chosen as a trade off between fitting the model to data ($K$ large) and limit the complexity of the model and
avoid overfitting ($K$ small). The best value is usually determined after performing cross-validation.

As mentioned above, the case $\LL(S)=0$ leads to a $K$-model. By choosing $\LLtrans(i,j)=\lambda$
for all $i\neq j$, $\LLmode(i)=\LLtrans(i,i)=0$, one penalizes mode transitions equally by $\lambda\geq 0$, where $\lambda\rightarrow\infty$ leads to regression of a single model on the data
(that is, $s_t\equiv s_0$), while $\lambda\rightarrow 0$ leads again to a $K$-model. Note that choosing the same constant $\lambda$ for
all transitions makes the fitting problem exhibit multiple solutions, as indexes $i$, $j$ can be
arbitrarily permuted. The mode loss $\LLmode$ can be used to break such symmetries. For example, smaller values for $s_t$ will be preferred
by making $\LLmode(i)<\LLmode(j)$ for $i<j$.
The shape of the increasing finite sequence $\{\LLmode(i)\}_{i=1}^K$ can be used to 
reduce the number of possible modes: larger increasing values of $\LLmode(i)$ will
discourage the use of an increasing number of modes.

The initial mode cost $\LLinit$ summarizes prior knowledge about the initial mode $s_0$. 
For example, $\LLinit(s_0)\equiv 0$ if no prior information on $s_0$ is available. On the contrary, if the initial mode $s_0$ is known and say equal to $j$, then $\LLinit(s_0)=0$ for $s_0=j$ and $+\infty$ otherwise.

Next Section~\ref{sec:stoch-interpret} suggests criteria for choosing $\LL$
in case statistical assumptions about the underlying process that generates $s_t$
are available. Alternative
criteria are discussed in Section~\ref{sec:auto-LL} for choosing $\LL$ directly from
the training data.

\section{Statistical interpretations}
\label{sec:stoch-interpret}

Let $Y=(y_1,\ldots,y_T)$, $X=(x_1,\ldots,x_T)$, $S=(s_0,\ldots,s_T)$, $\Theta = (\theta_1,\ldots,\theta_K)$. We provide a statistical interpretation
of the loss functions for the special case in which
the following modeling assumptions are satisfied:
\begin{itemize}
\item [A1.] The mode sequence $S$, the model parameters $\Theta$ and the input data $X$ are statistically independent, i.e.,
\[
p(S|X,\Theta) = p(S), \ \ \ p(\Theta|S,X) = p(\Theta)
\]
\item [A2.] The conditional likelihood of $Y$  is given by 
\begin{equation*}
p(Y|X,S,\Theta)=\prod_{t=1}^T p(y_t|X,S,\Theta)=\prod_{t=1}^T p(y_t|x_t,\theta_{s_t})
\end{equation*}
 where  $p(y_t|x_t,\theta_{s_t})$ is the likelihood of the outcome $y_t$ given $x_t$ and $\theta_{s_t}$;
\item [A3.] The priors on the model parameters $\theta_1,\ldots,\theta_K$ are all equal to $p(\theta)$, i.e., 
\begin{equation*}
p(\theta_1)= \cdots = p(\theta_K) = p(\theta)
\end{equation*}
and the model parameters are statistically independent, i.e.,
 \begin{equation*}
p(\Theta)=\prod_{k=1}^K p(\theta_k) 
\end{equation*}
\item [A4.] The probability of being in mode $s_t$ given
$s_0,\ldots,s_{t-1}$ is $p(s_t|s_t-1)=\pi_{s_t,s_{t-1}}$ (Markov property);  
\item [A5.] The initial mode $s_0$ has probability $p(s_0)=\pi_{s_0}$.
\end{itemize}

\begin{proposition}
Let Assumptions A1-A5 be satisfied and define
\begin{subequations} 
\beqar
    \ell(x_t,y_t,\theta_{s_t})&=&-\log p(y_t|x_t,\theta_{s_t})\label{eq:prob-y}\\
r(\theta_k)&=&-\log p(\theta_k)\label{eq:prob-rtheta}\\
\LLtrans(s_t,s_{t-1})&=&-\log \pi_{s_t,s_{t-1}}\label{eq:prob-LLtrans}\\
\LLinit(s_0)&=&-\log \pi_{s_0}\label{eq:prob-LLinit}\\
\LLmode(s_t)&=&0.
\eeqar%
\label{eq:prob-tot}%
\end{subequations}
Then minimizing $J(X,Y,\Theta,S)$ as defined in~\eqref{eq:criterion}--\eqref{eq:prob-tot}
with respect to $\Theta$ and $S$
is equivalent to maximizing the joint likelihood $p(Y,S,\Theta|X)$.
\label{prop:stat}
\end{proposition}
\emph{Proof}.
Because of the Markov property (Assumption~A4), the likelihood of the mode sequence $S$ is
\begin{equation} \label{eq:pS}
    p(S)=p(s_0)\prod_{t=1}^Tp(s_t|s_{t-1}).
\end{equation}
From~\eqref{eq:pS} and Assumptions~A1-A3, we have:
\beqarno
    p(Y,S,\Theta|X)&=&p(\Theta|X)p(Y,S|X,\Theta)\\&=&p(\Theta|X)p(S|X,\Theta)p(Y|S,X,\Theta)\\
                             &=&p(\Theta)p(S)p(Y|S,X,\Theta)\\
&=&\prod_{k=1}^Kp(\theta_k)p(s_0)\prod_{t=1}^Tp(s_t|s_{t-1})p(y_t|x_t,\theta_{s_t})
\eeqarno
whose logarithm is
\begin{equation}
    \begin{split}
    \log p(Y,S,\Theta|X)=&\sum_{k=1}^K\log p(\theta_k)+\log p(s_0)\\
&\hspace*{-1cm}+\sum_{t=1}^T\log p(s_t|s_{t-1})+\sum_{t=1}^T \log p(y_t|x_t,\theta_{s_t}).
    \end{split}
\label{eq:probability}
\end{equation}
By defining the loss functions $\ell$, $r$, $\LLtrans$, $\LLinit$, and $\LLmode$
as in~\eqref{eq:prob-tot}, the minimization of the fitting 
objective $J(X,Y,\Theta,S)$ as in~\eqref{eq:criterion}--\eqref{eq:cost-form}
with respect to $\Theta$ and $S$ is equivalent to maximizing the logarithm of the joint likelihood $p(Y,S,\Theta|X)$, and therefore $p(Y,S,\Theta|X)$.
\hfill$\blacksquare$

The following proposition provides an inverse result, namely a statistical interpretation of 
minimizing a given generic $J(X,Y,\Theta,S)$ defined as in~\eqref{eq:cost-form}. 
\begin{proposition}
Define the probability density functions
	\begin{subequations}
		\beqar
		p(y_t|x_t,\theta_{s_t})&=&\!\frac{e^{-\ell(x_t,y_t,\theta_{s_t})}}{\nu(\theta_{s_t},x_t)}\\
		p(S,\Theta|X)&=&\!\frac{\nu(S,\Theta,X) e^{-\LL(S)-r(\Theta) }  }{\!\!\sum\limits_{\bar S\in K^{T+1}} \displaystyle{\!\!\int\limits_{\rr^{d\times K}}
				\!\!\nu(\bar S,\Theta,X)e^{-\LL(S)-r(\Theta)}d\Theta}} \label{eq:prop2:pST}
        \eeqar%
	\end{subequations}
	where
	\begin{subequations}
		\beqar
		\nu(\theta_{s_t},x_t)&=&\int_{\YY} e^{-\ell(x_t,y,\theta_{s_t})}dy\label{eq:beta-1}\\
		\nu(S,\Theta,X)&=&\prod_{t=1}^T\nu(\theta_{s_t},x_t)\label{eq:beta-12}
		\eeqar
		\label{eq:beta}%
	\end{subequations}	
and assume that the outputs $Y$ are conditionally independent given $(S,X,\Theta)$, i.e., $p(Y|S,X,\Theta)=\prod_{t=1}^T p(y_t|x_t,\theta_{s_t})$. Then the following identity holds
\begin{equation}
\argmin\limits_{S,\Theta}   J(X,Y,\Theta,S)   = \argmax\limits_{S,\Theta}\log p(Y,S,\Theta|X)
\label{eq:maxlog}
\end{equation}
	\label{prop:stat-inv}
\end{proposition}
\emph{Proof}. Since
\begin{equation} \label{eq:prop2-P}
p(Y,S,\Theta|X) = p(Y|S,X,\Theta)p(S,\Theta|X)
\end{equation}
by substituting~\eqref{eq:beta} in~\eqref{eq:prop2-P} we get 
\begin{align} 
& p(Y,S,\Theta|X) = \nonumber  \\ 
& \frac{\prod_{t=1}^T e^{-\ell(x_t,y_t,\theta_{s_t})}}{\prod_{t=1}^T \nu(\theta_{s_t},x_t)} \frac{\nu(S,\Theta,X) e^{-\LL(S)-r(\Theta) }  }{\!\!\sum\limits_{\bar S\in K^{T+1}} \displaystyle{\!\!\int\limits_{\rr^{d\times K}}
		\!\!\nu(\bar S,\Theta,X)e^{-\LL(\bar S)-r(\Theta)}d\Theta}}    \nonumber   \\
 & =\frac{ e^{-\sum_{t=1}^T\ell(x_t,y_t,\theta_{s_t})  -\LL(S)-r(\Theta)   }}{\sum\limits_{\bar S\in K^{T+1}} \displaystyle{\!\!\int\limits_{\rr^{d\times K}}
		\!\!\nu(\bar S,\Theta,X)e^{-\LL(\bar S)-r(\Theta)}d\Theta}}    \label{eq:prop2-P2}  
\end{align}
As the denominator in~\eqref{eq:prop2-P2} does not depend on $S$ and $\Theta$, maximize $p(Y,S,\Theta|X)$ is equivalent to maximize
\begin{equation*}
 e^{-\sum_{t=1}^T\ell(x_t,y_t,\theta_{s_t})  -\LL(S)-r(\Theta)   },
\end{equation*} 
or, equivalently, to minimize
\begin{equation*}
\sum_{t=1}^T\ell(x_t,y_t,\theta_{s_t})  +\LL(S)+r(\Theta)
\end{equation*}
The identity~\eqref{eq:maxlog}  thus follows from the definition of $J(X,Y,\Theta,S)$ in~\eqref{eq:cost-form}. 
\hfill$\blacksquare$

The following corollary provides a set of probabilistic  interpretations   of the loss function $J(X,Y,\Theta,S)$,  some of which are well known in Bayesian estimation.

\begin{corollary}
	Let $\nu(\theta_{s_t},x_t)$ in~\eqref{eq:beta-1} be a constant. 
	Then the following statements hold:
	\begin{enumerate}
		\item The quadratic regularization $r(\Theta)=\rho \sum_{k=1}^K \|\theta_k\|_2^2$  corresponds to assuming a Gaussian prior on $\theta_k$, namely $p(\theta_k)=ce^{-\frac{\|\theta_k\|_2^2}{2\sigma_\theta^2}}$ with  $\sigma_\theta=\sqrt{\frac{1}{2\rho}}$.
		\item The quadratic penalty on the prediction error
		\begin{equation}
		\ell(x_t,y_t,\theta_{s_t})=c\|y_t-\theta_{s_t}'x_t\|_2^2   
		\label{eq:Gaussian-noise}
		\end{equation}
		correspond to assuming the probabilistic model of the output  $y_t\sim N(\theta_{s_t}'x_t,\sigma_y^2 I)$, with $\sigma_y=\sqrt{\frac{1}{2c}}$.
	\item Setting $\LLtrans=0$ is equivalent to assuming that the modes $s_t$ are i.i.d., with 
	\begin{equation*}s_t \sim p(s_t) = \frac{e^{-\LLmode(s_t)}}{\sum_{k=1}^Ke^{-\LLmode(k)}} \end{equation*}
	Furthermore, setting $\LL(S)=0$ corresponds to assuming that 
	$p(s_t)=\frac{1}{K}$ for all $t=0,\ldots,T$, while setting $\LLinit(s)=\LLmode(s)=s$, $s=1,\ldots,K$, corresponds to assuming
	$p(s)=\frac{(e-1)}{1-e^{-K}}e^{-s}$. 
	\item Under the assumption  $p(S|\Theta,X)=p(S)=p(s_0)\prod_{t=1}^Tp(s_t|s_{t-1})$, the case $\LLmode=\LLinit=0$ and $\LLtrans(i,j)=\lambda$ for $i\neq j$ and $0$ for $i=j$, 
	corresponds to assume that 
	\[
	p(s_0)\!=\!\frac{1}{K},\ \ p(s_t|s_{t-1})\!=\!\left\{\!
	\begin{array}{rll} \!\!
	\frac{e^{-\lambda}}{1+(K-1)e^{-\lambda}} & \mbox{if} & s_t\neq s_{t-1}\\
	\frac{1}{1+(K-1)e^{-\lambda}}& \mbox{if} & s_t= s_{t-1}
	\end{array}
	\right.
	\]
	\end{enumerate}
\end{corollary}   
 \emph{Proof}.  As $\nu(\theta_{s_t},x_t)$ does not depend on $\theta_{s_t}$ and $X$,  $p(S,\Theta|X)$ in~\eqref{eq:prop2:pST} can be written as $p(S,\Theta|X)= p(S)p(\Theta)$, where
	\begin{align} \label{eqn:prop-3-p}
	p(S)  =  \frac{ 
		\!\!e^{-\LL(S)}}{\sum\limits_{\bar S\in K^{T+1}}  \!\! 
		\!\!e^{-\LL(S)}}, \quad 
	p(\Theta) 	 = \frac{e^{-r(\Theta)}d\Theta}{\int\limits_{\rr^{d\times K}}
	 e^{-r(\Theta)}d\Theta} 
	\end{align}
The results follow straightforwardly from the above expressions of $p(S)$ and $p(\Theta)$ and the definition of $\LL(s)$ in~\eqref{eq:modeseqloss}. \hfill $\blacksquare$

\section{Algorithms}
\label{sec:algorithms}
We provide now algorithms for fitting a jump model to a given data set and to infer predictions $\hat y_t$, $\hat s_t$ from it.

\subsection{Model fitting}

Given a training sequence $X=(x_1,\ldots,x_T)$ of inputs and $Y=(y_1,\ldots,y_T)$ of outputs,
for fitting a jump $K$-model we need to attempt minimizing the cost $J(X,Y,\Theta,S)$ with respect to $\Theta$ and $S$. A simple algorithm to solve this problem is Algorithm~\ref{algo:JM-fit}, 
a coordinate descent algorithm that alternates minimization with respect to $\Theta$ and $S$. If $\ell$ and $r$ are convex functions, Step~\ref{algo1:Theta_fit}  can be solved globally (up to the desired precision) by standard convex programming~\cite{BV04}. Step~\ref{algo1:S_fit} can be solved to global optimality by standard discrete \emph{dynamic programming} (DP)~\cite{Bel57} 
with complexity $O\left(TK^2\right)$.  
This is achieved by computing the following matrices $M\in\rr^{K\times (T+1)}$
of costs and $U\in\KK\times \rr^{T}$ of indexes
\begin{subequations} \label{eq:DP-eq}
\begin{align}
    M(s,T)=&\LLmode(s)+\ell(x_T,y_T,\theta_s)\label{eq:DP-T}\\
    U_{s,t}=&\argmin_j\{M(j,t+1)+\LLtrans(j,s)\},\nonumber \\&t=1,\ldots,T-1\\
    M(s,t)=&\LLmode(s)+\ell(x_t,y_t,\theta_s)+
            M(U_{s,t},t+1)\nonumber\\&+\LLtrans(U_{s,t},s)\\
    M(s,0)=&\LLinit(s)+\min_j\{M(j,1)+\LLtrans(j,s)\}
\end{align}
backwards in time, and then reconstructing the minimum cost sequence $S$ 
forward in time by setting
\begin{align}
    s_0&=\argmin_j M(j,0)\\
    s_t&=U_{s_{t-1},t},\ t=1,\ldots,T.
\end{align}
\label{eq:DP}%
\end{subequations}%
Note that if the time order of operations in~\eqref{eq:DP} is reversed, the DP iterations~\eqref{eq:DP} become Viterbi algorithm~\cite[p. 264]{Rab89}:
\begin{subequations}
\begin{align}
    M(s,0)=&\LLinit(s)\\
    U_{s,t}=&\argmin_j\{M(j,t-1)+\LLtrans(j,s)\},\nonumber\\&t=1,\ldots,T\\
    M(s,t)=&\LLmode(s)+\ell(x_t,y_t,\theta_s)+M(U_{s,t},t-1)\nonumber\\&+\LLtrans(U_{s,t},s)
\end{align}
followed by the backwards iterations 
\begin{align}
    s_T&=\argmin_j M(j,T)\\
    s_t&=U_{s_{t+1},t},\ t=0,\ldots,T-1.
\end{align}
\label{eq:Viterbi}%
\end{subequations}%
Since at each iteration the cost $J(X,Y,\Theta,S)$ is non-increasing
and the number of sequences $S$ is finite, Algorithm~\ref{algo:JM-fit} always terminates in a finite number of steps, assuming that in case of multiple optima one selects
the optimizers in Steps~\ref{algo1:Theta_fit} and \ref{algo1:S_fit} according
to some predefined criterion. However, there is no guarantee
that the solution found is the global one, as it depends on the initial guess $S^0$. To improve the quality of the solution, we may run Algorithm~\ref{algo:JM-fit} $N$ times from different random initial sequences $S^0$ and select the best result. Our experience is that a small $N$, say $N=5$, is usually enough.

\begin{algorithm}[t]
\caption{Jump model fitting}
\label{algo:JM-fit}
~~\textbf{Input}: Training data set $X=(x_{1},\ldots,x_{T})$,
$Y=(y_1,\ldots,y_T)$,
number $K$ of models, initial mode sequence $S^0=\{s_0^0,\ldots,s^0_T\}$.
\vspace*{.1cm}\hrule\vspace*{.1cm}
\begin{enumerate}[label=\arabic*., ref=\theenumi{}]
\item \label{algo1:model_fit}
\textbf{iterate for} $k=1,\ldots$ 
\begin{enumerate}[label=\theenumi{}.\arabic*., ref=\theenumi{}.\theenumii{}]
\item \label{algo1:Theta_fit} $\Theta^{k}\leftarrow\argmin_\Theta \ell(X,Y,\Theta,S^{k-1})+r(\Theta)$;
\item[] \flushright\textsf{(model fitting)}\\[.5em]
\end{enumerate}
\begin{enumerate}[label=\theenumi{}.\arabic*., ref=\theenumi{}.\theenumii{}]
\addtocounter{enumii}{1}
\item \label{algo1:S_fit} $S^{k}\leftarrow\argmin_S \ell(X,Y,\Theta^k,S)+\LL(S)$;
\item[]\flushright\textsf{(mode sequence fitting)}
\end{enumerate}
\item \textbf{until $S^{k}=S^{k-1}$}.\label{algo1:stop} 
\end{enumerate}
\vspace*{.1cm}\hrule\vspace*{.1cm}
~~\textbf{Output}: Estimated model parameters $\Theta^\star=\Theta^k$ and mode sequence $S^\star=S^k$.
\end{algorithm}

During the execution of Algorithm~\ref{algo:JM-fit} it may happen that a mode $s$ does not appear in the sequence $S^{k-1}$. In this case, the \emph{fitting loss} $\ell(X,Y,\Theta,S^{k-1})$ does not depend on $\theta_s$, and the latter will be determined in Step~\ref{algo1:Theta_fit}  based only on the regularizer $r(\Theta)$.

In case $\LL(S)=0$, the ordering of the training data becomes irrelevant and
Algorithm~\ref{algo:JM-fit} reduces to fitting $K$ models to the data set.  
If in addition $\ell$ and $r$ are specified as in~\eqref{eq:kmeans} and
$Y=X$, Algorithm~\ref{algo:JM-fit} is the standard $K$-means algorithm,
where the starting sequence $S^0$ is the initial clustering of the data points $(x_1,\ldots,x_T)$, Step~\ref{algo1:Theta_fit} computes the collection $\Theta^{k}$ of cluster centroids at iteration $k$, and Step~\ref{algo1:S_fit} reassigns data points to clusters by updating their labels $s^k_t$. 

When again $\LL(S)=0$ and the mode loss in~\eqref{eq:pwa} is used
for getting a PWA model, the cost function minimized in Step~\ref{algo1:Theta_fit} of Algorithm~\ref{algo:JM-fit} is separable with respect to $\theta_{y,s}$, $\theta_{x,s}$. Then the minimization with respect to $\theta_{x,s}$ produces the piecewise linear separation function $\max_s\{\theta_{x,s}'\smallmat{x\\1}\}$ that defines the polyhedral partition of the input space~\cite{BPB16a}, while Step~\ref{algo1:S_fit} looks for the optimal latent variables $s_t$ that best trade off between
assigning the corresponding data point $x_t$ to the polyhedron
$\{x\in \XX: \theta_{x,s_t}'\smallmat{x\\1}\geq \theta_{x,j}'\smallmat{x\\1},\ \forall j\neq s_t, j\in\KK\}$ and matching the predicted output
$y_t\approx\theta_{y,s_t}'\smallmat{x_t\\1}$.

Finally, we remark that Algorithm~\ref{algo:JM-fit} is also applicable to the more
general case in which the mode loss $\LL$ also depends on $\Theta$, by simply replacing
Steps~\ref{algo1:Theta_fit} and~\ref{algo1:S_fit} with
\begin{subequations}
\begin{align}
\Theta^{k}&\leftarrow\argmin_\Theta \ell(X,Y,\Theta,S^{k-1})+r(\Theta)+\LL(S^{k-1},\Theta)\\
S^{k}&\leftarrow\argmin_S \ell(X,Y,\Theta^k,S)+\LL(S,\Theta^k). 
\end{align}
\label{eq:L(S,Theta)}%
\end{subequations}
This would cover the case in which $\LL$ contains parameters to be estimated.

\subsection{Inference}
\subsubsection{One-step ahead prediction}
\label{sec:ol-inference}
Assume that the model parameters $\Theta^\star$ have been estimated
and that new production data $\tilde X_t=(\tilde x_1,\ldots,\tilde x_t)$
and outputs $\tilde Y_{t-1}=(\tilde y_1,\ldots,\tilde y_{t-1})$
are given. Because of the structure of the mode loss function $\LL$ 
defined in~\eqref{eq:modeseqloss}, the estimates $\hat y_t$ and $\hat s_0,\ldots,\hat s_t$ do not depend on future inputs $\tilde x_j$ and modes $\hat s_j$ for $j>t$.

The same  fitting objective~\eqref{eq:criterion} can be used to estimate $\hat y_t$ and $\hat S_t=(\hat s_0,\ldots,\hat s_t)$,
\begin{equation}
\begin{split}
      (\hat y_t,\hat S_t)=&\argmin_{y,S_t} J_t(\tilde X_t,\tilde Y_{t-1},y,\Theta^\star,S_t)\\
        &\st\ y\in\YY_t
\label{eq:inference}
\end{split}
\end{equation}
where  $\YY_t\subseteq\YY$ is a possible additional output information set and
\begin{equation}
\begin{split}
    J_t(\tilde X_t,\tilde Y_{t-1},y,\Theta^\star,S_t)=\ell(\tilde x_t,y,\theta^\star_{s_t})+\sum_{j=1}^{t-1}\ell(\tilde x_j,\tilde y_j,\theta^\star_{s_j})\nonumber\\
+\LLinit(s_0)+\sum_{j=1}^t\LLmode(s_j)+ 
    \sum_{j=1}^t\LLtrans(s_j,s_{j-1}).
\end{split}
\label{eq:J-inference}
\end{equation}

Algorithm~\ref{algo:JM-inference} attempts at solving problem~\eqref{eq:inference} at every $t$ of interest. Step~\ref{algo:JM-inference-DP}
is solved again by the DP iterations~\eqref{eq:DP}
over the time span $[0,t]$, with the only difference that in~\eqref{eq:DP-T} we set the terminal penalty equal to $M(s,t)=\LLmode(s)+\min_y\{\ell(\tilde{x}_t,y,\theta_s)\}$, since the last output $y_t$ is determined later at Step~\ref{algo:JM-inference-yhat}.

Note that \emph{open-loop prediction}, that is the task of
predicting $\hat y_t$ and $\hat s_t$ without acquiring $\tilde Y_{t-1}$, can be simply obtained by replacing
$\tilde Y_{t-1}=(\tilde y_1,\ldots,\tilde y_{t-1})$ with
$\hat Y_{t-1}=(\hat y_1,\ldots,\hat y_{t-1})$. Arbitrary combinations
of one-step ahead and open-loop predictions are possible
to handle the more general case of intermittent 
output data availability.

\begin{algorithm}[t]
\caption{Inference}
\label{algo:JM-inference}
~~\textbf{Input}: Model set $\Theta^\star$, production data set $\tilde X_t=(\tilde x_1,\ldots,\tilde x_t)$, past outputs $\tilde Y_{t-1}=(\tilde y_1,\ldots,\tilde y_{t-1})$.
\vspace*{.1cm}\hrule\vspace*{.1cm}
\begin{enumerate}[label=\arabic*., ref=\theenumi{}]
\item \label{algo:JM-inference-DP} $\displaystyle{\hat S_t\leftarrow\argmin_{S_t} \left\{\LL(S_t)+\sum_{j=1}^{t-1}\ell(\tilde x_j,\tilde y_j,\theta^\star_{s_j})\right.}$
\item []\flushright$\displaystyle{\left.+\min_{y\in\YY_t}\ell(\tilde x_t,y,\theta^\star_{s_t})\right\}}$;
\end{enumerate}
\begin{enumerate}[label=\arabic*., ref=\theenumi{}]
\addtocounter{enumi}{1}
\item \label{algo:JM-inference-yhat} $\hat y_t\leftarrow\argmin_{y\in\YY_t} \ell(\tilde x_t,y,\theta^\star_{\hat s_{t}})$;
\end{enumerate}
\vspace*{.1cm}\hrule\vspace*{.1cm}
~~\textbf{Output}: Estimated output $\hat y_t$ and mode sequence $\hat S_t$.
\end{algorithm}

\subsubsection{Recursive inference}
When $\LLtrans=0$, problem~\eqref{eq:inference} becomes
completely separable and simplifies to
\begin{equation}
    (\hat y_t,\hat s_t)=\argmin_{y,s}
    \ell(\tilde x_t,y,\theta^\star_{s})+\LLmode(s)\quad \st\ y\in\YY_t.
\label{eq:predictor2}
\end{equation}
For example, in the case of $K$-means~\eqref{eq:kmeans} ($\LL(s)=0$), the
estimate obtained by~\eqref{eq:predictor2} is given by~\eqref{eq:styt-kmeans}.

When the mode transition loss function $\LLtrans\neq 0$, the simplification in~\eqref{eq:predictor2} does not hold anymore. Nonetheless,
an incremental version of~\eqref{eq:inference} can be still derived as described in  Algorithm~\ref{algo:JM-inference_rec}, where   $\LL_{t}:\KK\to\rr$ is the \emph{arrival cost} recursively computed by the algorithm from the  initial condition $\LL_{0}(s_0)= \LLinit(s_0)$, for all $s_0 \in \KK$.
\begin{algorithm}[t]
	\caption{Recursive inference}
	\label{algo:JM-inference_rec}
	~~\textbf{Input}: Model $\Theta^\star$, current input $\tilde x_t$, past input/output pair $(\tilde x_{t-1},\tilde y_{t-1})$, arrival cost $\LL_{t-1}$.
	\vspace*{.1cm}\hrule\vspace*{.1cm}
	\begin{enumerate}[label=\arabic*., ref=\theenumi{}]
		\begin{subequations} \label{eq:incremental-predictor-eq}
		\item \textbf{Update} \label{algo:JM-inference-DP_rec} 
					\beqar
		\LL_{t}(s_{t})&\leftarrow &\LLmode(s_{t})+
		\min_{s_{t-1}}\left\{\ell(\tilde x_{t-1},\tilde y_{t-1},\theta_{s_{t-1}})\right.\nonumber\\
		&&\left.+\LL_{t-1}(s_{t-1})+\LLtrans(s_{t},s_{t-1})\right\}
		\label{eq:arrival-cost}
		\eeqar
		\item \textbf{Compute}	
		\begin{equation}
		(\hat y_t,\hat s_t) \leftarrow \argmin_{y,s} 
		\ell(\tilde x_t,y,\theta_{s})+\LL_{t}(s)\quad \st\ y\in\YY_t
		\label{eq:incremental-predictor}
		\end{equation} 
		\end{subequations}
	\end{enumerate}
	\vspace*{.1cm}\hrule\vspace*{.1cm}
	~~\textbf{Output}: Estimated output $\hat y_t$ and mode $\hat s_t$, updated arrival cost  $\LL_t$.
\end{algorithm} 
Clearly, while producing exactly the same results,
the formulation in Algorithm~\ref{algo:JM-inference_rec} is much more
efficient than Algorithm~\ref{algo:JM-inference}, 
as the number of computations does not increase with $t$ and thus can be used  for online inference.

\subsubsection{Smoothing} \label{sec:smoothing}
The same approach described in Section~\ref{sec:ol-inference}
can be generalized to other inference tasks than one-step ahead or open-loop prediction, such as \emph{smoothing}. Assume $\tilde y_k$ is only known at steps $k\in\TT_t\subseteq\{1,\ldots,t\}$. Steps~\ref{algo:JM-inference-DP}--\ref{algo:JM-inference-yhat} of Algorithm~\ref{algo:JM-inference} are
replaced by 
\begin{subequations} \label{eq:smoothing}
\begin{align}
\hat S_t&\leftarrow\argmin_{S_t} \hspace*{-.0cm}\Big\{\LL(S_t)+\sum_{j\in\TT_t}\ell(\tilde x_j,\tilde y_j,\theta^\star_{s_j})\Big. \nonumber \\
&\hspace{2.5cm} \Big.+\sum_{j\in\bar{\TT_t}}\min_{y_j\in\YY_j}\ell(\tilde x_j,y_j,\theta^\star_{s_j})\Big\} \label{eq:smoothing-v1}
\\
y_j&\leftarrow\argmin_{y\in\YY_j} \ell(\tilde x_j, y,\theta^\star_{\hat s_j}),\ \forall j\in\bar\TT_t
\end{align} 
\end{subequations}
where $\bar\TT_t=\{1,\ldots,t\}\setminus\TT_t$. Note that  complexity of the inner minimization in~\eqref{eq:smoothing-v1} depends on the shape of the loss function $\ell$. In the 
quadratic case, the minimum can be expressed analytically.

\subsubsection{Pure mode estimation} \label{sec:pure-mode-est}
In case we are interested in estimating only the latent mode $\hat s_t$ 
given $\tilde x_1,\ldots,\tilde x_t$, $\tilde y_1,\ldots,\tilde y_{t-1}$
and also $\tilde y_t$, 
we can keep using~\eqref{eq:incremental-predictor-eq}  
by simply changing~\eqref{eq:incremental-predictor} to
\begin{equation}
    \hat s_t=\argmin_{s} 
    \ell(\tilde x_t,\tilde y_t,\theta_{s})+\LL_{t}(s)
\label{eq:incremental-mode-predictor}
\end{equation}
This allows reconstructing the mode sequence $\hat s_1,\ldots,\hat s_{\tilde T}$ recursively
from the available data set, which may be useful for example to detect changes 
in the relation between the input $\tilde x_t$ and the output $\tilde y_t$.

\subsection{Relation with Hidden Markov Models} \label{eqn:secHMM}
Jump models have several common features with hidden Markov models (HMMs) \cite{Rab89}. 
First, both models consider the presence of discrete latent states $s_t$.
While HMMs assume that the sequence $S$ of such states satisfy the Markov property  
\[
	p(s_t|s_{t-1},\ldots,s_0)=p(s_t|s_{t-1})
\]
in jump models the particular form chosen in~\eqref{eq:modeseqloss}
for the mode sequence loss $\LL$ makes estimating $\hat s_t$ incrementally as in~\eqref{eq:incremental-predictor-eq} possible.

Second, in HMMs the observed outputs are such that
\[
	p(y_t|x_t,\ldots,x_1,y_{t-1},\dots,y_1,s_t,\ldots,s_0)=p(y_t|x_t,s_{t}).
\]
Similarly, in jump models $\hat y_t$ is a unique function of a given pair $(x_t,s_t)$,
as~\eqref{eq:incremental-predictor} becomes
\[
   \hat y_t=\argmin_{y\in\YY_t} \ell(x_t,y,\theta_{s_t}).
\]
Indeed, an HMM is a special case of a jump model.
Consider the case in which the output observation $y_t$ is discrete, that is  $\YY=\{1,\ldots,L\}$. An HMM is characterized by the set
of discrete probabilities
\begin{subequations}
	\begin{align}
	p(s_{t+1}=i|s_t=j)&=\pi_{i,j},\quad i,j\in\KK\\
	p(s_0)&=\pi_{s_0}\\
	p(y_t=v|s_t=j)&=\beta_{j,v},\ v\in \YY.
	\end{align}
\end{subequations}
Let us set $x_t=1$, $\theta_s=s$, and define the loss function $\ell$ as
\begin{equation}
\ell(x,y,\theta_s)=-\log(\beta_{\theta_s,y}).
\label{eq:ell-HMM}
\end{equation}
Similarly to~\eqref{eq:prob-tot}, by also setting $r(\theta)=0$, $\LLtrans(s_t,s_{t-1})=-\log \pi_{s_t,s_{t-1}}$, $\LLinit(s_0)=-\log \pi_{s_0}$, and $\LLmode(s)=0$,
the jump model defined by the inference rule~\eqref{eq:inference}--\eqref{eq:J-inference} 
returns the mode sequence $\hat S_t$ that best matches
the observed sequence $\tilde Y_t$ of outputs 
and that sets the output $\hat y_t$
equal to the value $v\in \YY$ that maximizes the probability 
$\beta_{\hat s_t,v}$. An extension to HMMs with continuous observation densities can be obtained by properly redefining the loss function $\ell$ in~\eqref{eq:ell-HMM}.

In case the probabilities $\beta_{s,y}$ are not given, but rather must be estimated from a training data set, 
we can set instead $\theta_s=[\beta_{s,1}\ \ldots\ \beta_{s,L}]'$ 
along with the loss function $\ell$
\begin{equation}
\ell(y,\theta_s)=-\log(e_y'\theta_{s})
\label{eq:ell-HMM2}
\end{equation}
where $e_y$ is the $y$th column of the identity matrix of size $L$. 
If the initial probability distribution $\pi_{s_0}$ and the state transition probabilities $\pi_{i,j}$ are unknown, they can be estimated by minimizing $J(X,Y,\Theta,S)$ in~\eqref{eq:criterion} with $\LLtrans(s_t,s_{t-1})$ and $\LLinit(s_0)$ as in~\eqref{eq:prob-LLtrans} and~\eqref{eq:prob-LLinit}, respectively. This implies that the unknown model parameter $\Theta$ should also include $\pi_{s_0}$ and  $\pi_{i,j}$, leading
to the general case of having the mode sequence loss $\LL$ also dependent on $\Theta$ as in~\eqref{eq:L(S,Theta)}.

The well-known \emph{Expectation-Maximization} (EM) algorithm~\cite{DLR77} determines
the parameters of an HMM by maximizing the log-likelihood
\begin{equation*} \label{eq:LLHMM}
L_{\rm HMM}(\Theta|X,Y) =  \log p(Y|X,\Theta)  \!\! =  \! \log \!\!  \! \sum_{S\in\KK^{T+1}}  \!\! \!\! p(Y,S|X,\Theta)
\end{equation*}
with respect to $\Theta$. Instead, as shown by Proposition~\ref{prop:stat},
our approach maximizes $\log p(Y,S,\Theta|X)$ with respect to $\Theta$ and $S$.

The case of HMMs in which the observations $y$ are a mode-dependent linear function of $x$
rather than discrete has been dealt with for example in~\cite{Fri94}, under the assumption that such a linear relation between input and output samples is perturbed by Gaussian noise. This is a special case of our jump model framework, obtained by setting $\ell$ as in~\eqref{eq:Gaussian-noise}, $\LLtrans$ as in~\eqref{eq:prob-LLtrans}, $\LLinit$ as in~\eqref{eq:prob-LLinit}, $\LLmode(s)=0$, and $r(\theta)=0$. The training algorithm described in~\cite{Fri94}, however, completely relies on the probabilistic assumptions made about the normal distribution of output noise and the Markovian nature of mode transitions.

In conclusion, jump models are more descriptive than HMMs. The sequence of modes may not be generated by a Markov chain, such as in the case of PWA models~\eqref{eq:pwa} and~\eqref{eq:pwainference}, where the mode $s_t$ is a deterministic function of $x_t$. In addition, the loss and mode loss functions can have rather arbitrary shapes. For example, we may choose $\ell(x,y,\theta_s)$ as the Huber function of $y-\theta_s'x$ for robust regression, which is still a convex loss.

\subsection{Selecting the mode sequence loss from data}
\label{sec:auto-LL}
Selecting the right mode sequence loss $\LL$ may not be obvious
and require several attempts that involve fitting and cross-validation.
A simple approach to choose $\LL$ directly from the 
training data is to update the mode loss function $\LL$ after executing
Algorithm~\ref{algo1:model_fit} based on the best sequence $S^\star$
found so far, and run Algorithm~\ref{algo1:model_fit} again, executing 
the algorithm $N$ times in total.

Assuming $\LLmode=0$ and given a set of relative weights $\tau_0,\tau_1,\ldots,\tau_K$,  we update $\LLtrans$, $\LLinit$ from one run of Algorithm~\ref{algo1:model_fit}
to another by setting
\begin{subequations}
\beqar
    \mu_j&\leftarrow&\frac{\card \{t\in\{1,\ldots,T\}:\ s^\star_{t-1}=j\}}{T}\\
    \mu_{ij}&\leftarrow&\frac{\card \{t\in\{1,\ldots,T\}:\ s_{t}^\star=i,\ s^\star_{t-1}=j\}}{T}\nonumber\\\\
    \LLtrans(i,j)&\leftarrow&-\tau_i\frac{\log\left(\frac{\mu_{ij}}{\mu_j}\right)}{\sum_{j=1}^{K}\log\left(\frac{\mu_{ij}}{\mu_j}\right)}\\
    & & i,j=1,\ldots,K \nonumber \\
    \mu_j^0&\leftarrow&\frac{\card \{t\in\{0,\ldots,T\}:\ s^\star_{t}=j\}}{T}\\
    \LLinit(j)&\leftarrow&-\tau_0\frac{\log\left(\mu^0_j\right)}{\sum_{j=1}^{K}\log\left( \mu^0_j\right)} 
\eeqar%
\label{eq:a-posteriori-mode-loss}%
\end{subequations} %
where $\card$ denotes the cardinality (number of elements) of a set and $S^\star=(s^\star_{0},\ldots,s^\star_{T})$. The choice in~\eqref{eq:a-posteriori-mode-loss} preserves   the relative weight between the losses $\LL$, $\ell$, and  $r$, as
\begin{eqnarray*}
\tau_i&=& \sum_{j=1}^{K}\LLtrans(i,j),\quad  i=1,\ldots,K\\
\tau_0&=&\sum_{j=1}^{K}\LLinit(j) 
\end{eqnarray*}
remains the same each time $\LLinit(j)$ and $\LLtrans(i,j)$ are updated as in~\eqref{eq:a-posteriori-mode-loss}.
Choosing $\LL$ as in~\eqref{eq:a-posteriori-mode-loss} is motivated
by the statistical interpretation~\eqref{eq:prob-LLtrans}--\eqref{eq:prob-LLinit}
and used routinely for estimating state probabilities in HMMs~\cite{Rab89}.
Clearly,~\eqref{eq:a-posteriori-mode-loss} are well defined only if $\mu_{ij}$, $\mu_j$, $\mu^0_j>0$
for all $i,j=1,\ldots,K$. If the latter condition is not satisfied, one may consider adding the following \emph{Laplace smoothing}~\cite[Ch. 13]{manning2008introduction}:
\begin{subequations}
\beqar
    \mu_j&\leftarrow&\frac{1+\card \{t\in\{1,\ldots,T\}:\ s^\star_{t-1}=j\}}{T+K}\\
    \mu_{ij}&\leftarrow&\frac{1+\card \{t\in\{1,\ldots,T\}:\ s_{t}^\star=i,\ s^\star_{t-1}=j\}}{T+K^2}\\
    \mu_j^0&\leftarrow&\frac{1+\card \{t\in\{0,\ldots,T\}:\ s^\star_{t}=j\}}{T+K}
\eeqar%
\end{subequations}
when estimating $\mu_j$, $\mu_{ij}$ and $\mu_j^0$.

Computing $\LL$ according to~\eqref{eq:a-posteriori-mode-loss} after the training step
has been found especially useful for improving the quality of inference, both when using~\eqref{eq:J-inference} or~\eqref{eq:incremental-predictor}.

\section{Examples}
\label{sec:examples}
We test the algorithms proposed in the previous sections
on various problems of regression and classification using jump 
models. In all the examples, convex optimization methods are used to solve the problem at Step~\ref{algo1:Theta_fit} of Algorithm~\ref{algo:JM-fit}, while dynamic programming is used to compute the global optimum at Step~\ref{algo1:S_fit}.  As the DP computation also provides the optimal cost 
$V^k\triangleq J(X,Y,\Theta^k,S^k)$, when running the tests we replace the termination criterion in Step~\ref{algo1:stop} with 
\begin{equation}
V^{k-1}-V^{k}\leq \epsilon_V    
\label{eq:Vtol}
\end{equation}
where $\epsilon_V$ is a small tolerance. In all the examples we set $\epsilon_V=10^{-8}$.

Furthermore, after the end of the training step, the loss $\LL$ is updated as in~\eqref{eq:a-posteriori-mode-loss} before making inference.

All tests were run on a MacBook Pro 3~GHz-Intel i7 in MATLAB R2016b.
The test code is available for download at \url{http://cse.lab.imtlucca.it/~bemporad/jump_models/}.

\subsection{Jump linear model regression}
We consider a dataset of $T=10000$ training data 
and $\tilde T=10000$ production data generated
by the following jump linear model with $K=3$ modes
\[
    y_t=\theta_{s_t}x_t+\zeta_t
\]
with $y_t\in\rr$, $x_t\in\rr^{20}$, $x_{t,i} \sim \NN(0,1)$ for all $i=1,\ldots,20$,  $\zeta_t\sim\NN(0,\sigma_y^2)$. The coefficients
of the parameter vectors $\theta_i$ are randomly selected
from the normal distribution $\NN(0,1)$.
The true mode $s_t$
has probability $\pi=5\%$ of being different from $s_{t-1}$, starting from $s_0=1$.

We consider the loss functions
\beqarno
    \ell(x_t,y_t,\theta_{s_t})&=&\|y_t-\theta_{s_t}'x_t\|_2^2\\
r(\theta_k)&=&10^{-5}\|\theta_k\|_2^2\\
\LLtrans(s_t,s_{t-1})&=&\left\{\begin{array}{lll}
-\tau\log(1-(K-1)\pi) &\mbox{if} & s_t=s_{t-1}\\
-\tau\log \pi&\mbox{if} & s_t\neq s_{t-1}\\
\end{array}\right.\\
\LLinit(s_0)&=&0\\
\LLmode(s_t)&=&0
\eeqarno%
where $\tau$ is treated as a hyper-parameter to be tuned.
Algorithm~\ref{algo:JM-fit}  is executed $N=5$ times from different
random initial guesses. Each execution is limited to $k_{\rm max}=1000$ iterations.

\begin{figure}
\includegraphics[width=\hsize]{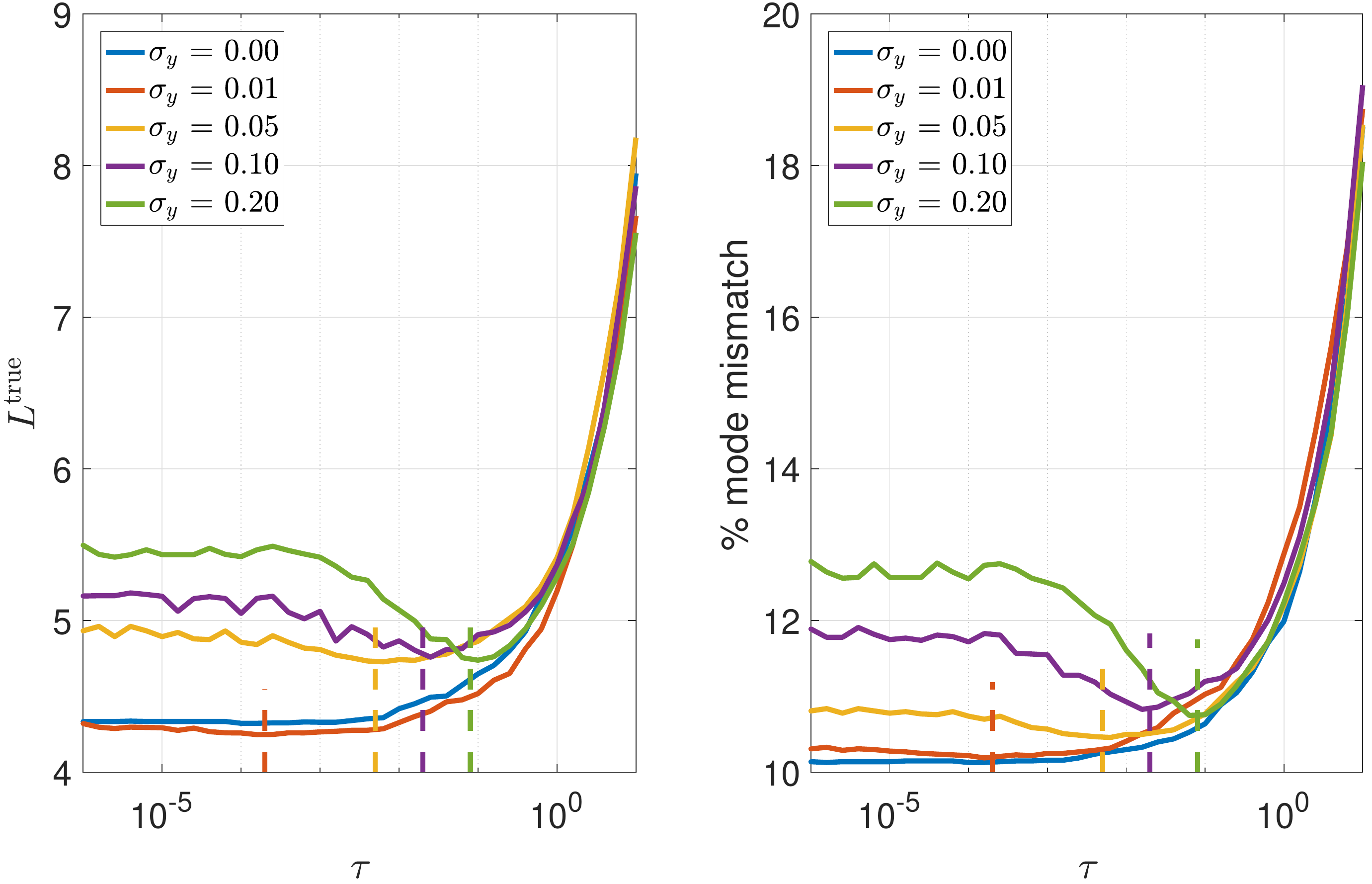}
\caption{Jump linear model fit and validation using recursive one-step ahead prediction: true loss $\tilde L^{\rm true}$ (left)
and mode mismatch $\ell_s^{\rm true}$ (right), 
optimal theoretical value $\tau^\star=2\sigma_y^2$ (dashed line)} \label{fig:jm-linear}
\end{figure}

We run Algorithm~\ref{algo:JM-fit} on the training data 
for different magnitudes $\sigma_y$ of output noise and values of the hyper-parameter
$\tau$. The resulting model coefficients $\Theta^\star$ are then
used in Algorithm~\ref{algo:JM-inference_rec} for recursive inference
on the production data. For assessing the quality of inference 
we use the true loss $L^{\rm true}$ defined in~\eqref{eq:trueloss}  with $ \ell^{\rm true}(\hat y_t,\tilde y_t)   = \|\hat y_t-\tilde y_t\|_2^2$.  
In addition, assuming the latent modes $\tilde s_t$ are available only for validation
purposes, we consider the following mode-mismatch figure
\begin{equation}
    \ell_s^{\rm true}=\frac{100}{\tilde T}\sum_{t=1}^{\tilde T}\delta_{\hat s_t,\tilde s_t}
\label{eq:mode-mismatch}
\end{equation}
where $\delta_{i,j}$ is the Kronecker delta function. The 
results are summarized in Figure~\ref{fig:jm-linear}. 

By recalling~\eqref{eq:Gaussian-noise} and~\eqref{eq:prob-LLtrans},
in order to minimize $-\log p(y_t|x_t,\theta)-\log\pi_{s_t,s_{t-1}}$
one should set $\ell(x_t,y_t,\theta_{s_t})=\frac{1}{2\sigma_y^2}$
and $\LLtrans(s_t,s_{t-1})=-\log \pi_{s_t,s_{t-1}}$, 
or equivalently $\ell(x_t,y_t,\theta_{s_t})=1$, 
$\LLtrans(s_t,s_{t-1})=-\tau^\star\log \pi_{s_t,s_{t-1}}$ with 
$\tau^\star=2\sigma_y^2$. Figure~\ref{fig:jm-linear} also reports
the value of $\tau^\star$ (dashed line) corresponding to different
values of $\sigma_y$. As expected, the best value for $\tau$ obtained
by cross validation, corresponding to the minimum of the plotted curves,
corresponds to the theoretical one $\tau^\star$ that would be obtained
if $\sigma_y$ were known. For large values of $\tau$ the percentage
of mode mismatch becomes close to $\frac{K-1}{K}\approx 66\%$ (not
shown in the figure), 
that is the value one gets when the mode $\hat s_t$ is assigned randomly. 
The average CPU time for executing Algorithm~\ref{algo:JM-fit} is $342$~ms, with 
the longest execution requiring $93$ iterations. Algorithm~\ref{algo:JM-inference_rec} 
requires $0.89~\mu$s per data point on average to make one-step ahead inference.

\begin{figure}
	\includegraphics[width=\hsize]{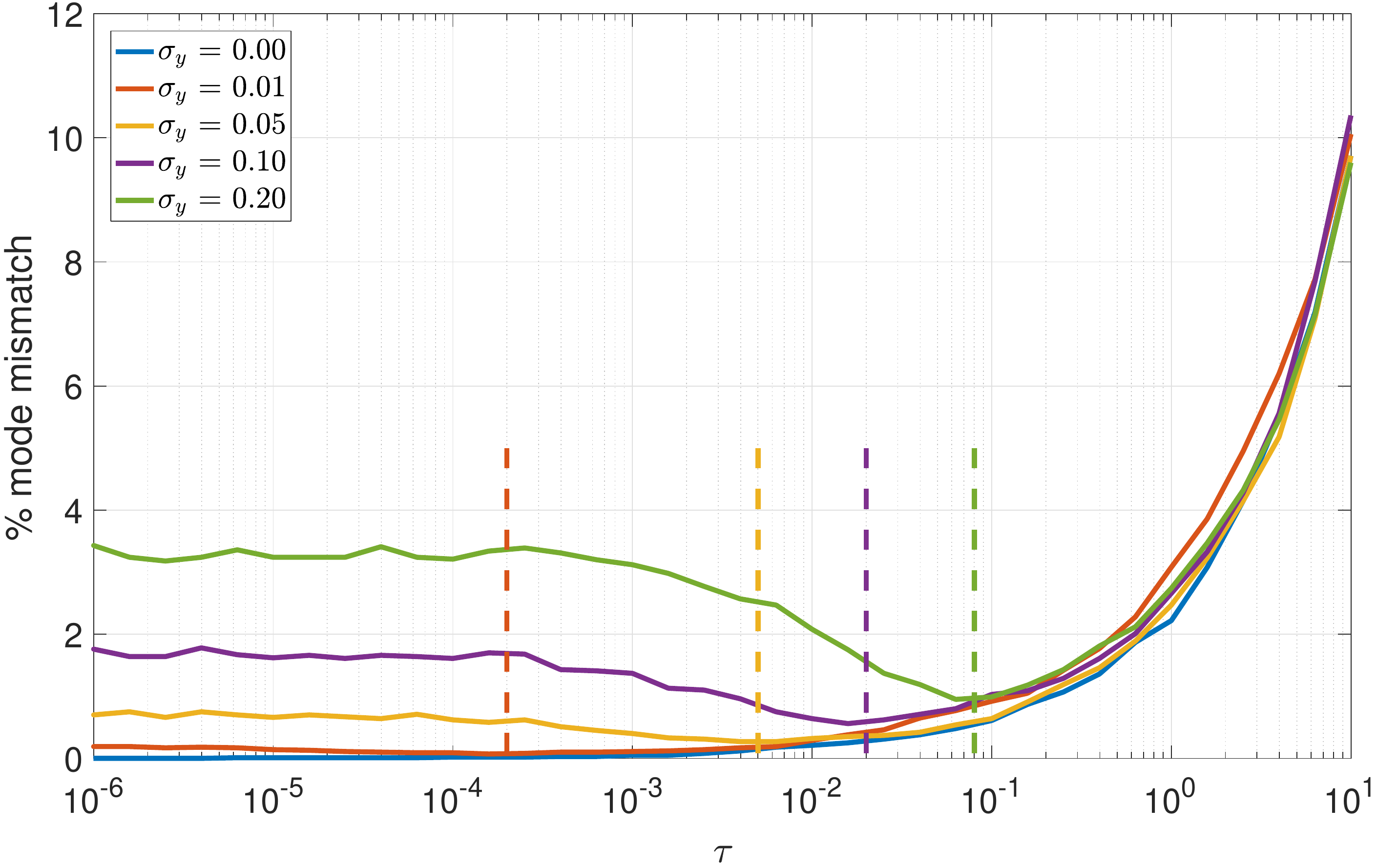}
	\caption{Jump linear model fit and validation using pure mode estimation: mode mismatch $\ell_s^{\rm true}$ (right), optimal theoretical value $\tau^\star=2\sigma_y^2$ (dashed line)} \label{fig:jm-linear2}
\end{figure}

Figure~\ref{fig:jm-linear2} shows the percentage of misclassified modes when
pure mode estimation, as presented in Section~\ref{sec:pure-mode-est}, is employed
instead of one-step ahead prediction. In this case, the latent mode $\hat s_t$ is reconstructed based not only on the observations  $\tilde x_1,\ldots,\tilde x_t$, $\tilde y_1,\ldots, \tilde y_{t-1}$ but also $\tilde y_t$, using Algorithm~\ref{algo:JM-inference-DP_rec}
with \eqref{eq:incremental-predictor} replaced by~\eqref{eq:incremental-mode-predictor}. 
As expected, compared to Figure~\ref{fig:jm-linear}, taking into account the current observation $\tilde y_t$ in estimating $\hat s_t$ reduces the number of misclassified modes.

Finally, the \emph{Expectation Maximization}  algorithm for HMM regression in~\cite{Fri94} is implemented and  compared with our method, with the hyper-parameter $\tau$ chosen,
for each different $\sigma_{y}$, as the best value observed in cross-validation. 
In EM the sequence
of latent modes is inferred in a batch way from the production dataset by using Viterbi algorithm~\cite{viterbi2010error}. In our approach, the mode sequence is estimated using  Algorithm~\ref{algo:JM-inference-DP_rec}
with \eqref{eq:incremental-predictor} replaced by~\eqref{eq:incremental-mode-predictor}.
Table~\ref{Tab:ex1} summarizes the results of the comparison, showing that our approach
provides a slightly better, although very similar, mode mismatch figure $\ell_s^{\rm true}$~\eqref{eq:mode-mismatch}.

\begin{table}[!tb]
\caption{Jump linear model validation, smoothing results: mode mismatch $\ell_s^{\rm true}$ achieved by the Expectation Maximization (EM) algorithm for HMM regression~\cite{Fri94} and by the approach discussed in this paper (Algorithms~\ref{algo:JM-fit} and~\ref{algo:JM-inference}).} \label{Tab:ex1}
		\begin{center}
\begin{tabular}{|c|C{2cm}|C{2cm}|}
		\cline{2-3}
		\multicolumn{1}{c|}{}  & \multicolumn{2}{c|}{$\ell_s^{\rm true}~\%$}\\
		\cline{2-3}
		\multicolumn{1}{c|}{} 
		&  EM 
		& \hspace{-0.2cm} Algorithms~\ref{algo:JM-fit}-\ref{algo:JM-inference} \hspace{-0.2cm}
		\\
		\hline
		$\sigma_{y}=0.00$ & 0.00 & 0.00\\[-.3em]
		$\sigma_{y}=0.01$ & 0.12 & 0.06\\[-.3em]
		$\sigma_{y}=0.05$  & 0.40 & 0.23\\[-.3em]
		$\sigma_{y}=0.10$ & 1.24 & 0.59\\[-.3em]
		$\sigma_{y}=0.20$  & 1.84 & 0.88\\
		\hline
\end{tabular}
\end{center}
\end{table}

\subsection{Jump binary classification}
We consider $T=10000$ training data 
and $\tilde T=10000$ production data generated
by the following jump linear model with $K=3$ modes
\[
    y_t=\sign(\theta_{s_t}x_t+\zeta_t)
\]
with 
\[
    \matrice{ccc}{\theta_1& \theta_2&\theta_3}=\smallmat{ -1 & -1 & -1\\
    1.1812 &  -0.5587  &  0.8003\\
   -0.7585 &   0.1784  & -1.5094\\
   -1.1096 &  -0.1969  &  0.8759\\
   -0.8456 &   0.5864  & -0.2428\\
   -0.5727 &   0.8759  &  0.6037\\
   -0.5587 &  -0.2428  &  1.7813\\
    0.1784 &   0.1668  &  1.7737}
\]
and $y_t\in\{-1,1\}$, $x_t\in\rr^8$, $x_{t,i} \sim \NN(0,\sigma_x^2)$ for all $i=1,\ldots,8$
with $\sigma_x=10$,  $\zeta_t\sim\NN(0,\sigma_y^2)$, $\sigma_y=0.1$. The true mode
$s_t$ changes every $500$ samples during the generation of the data, covering all modes.

\begin{figure}
\includegraphics[width=\hsize]{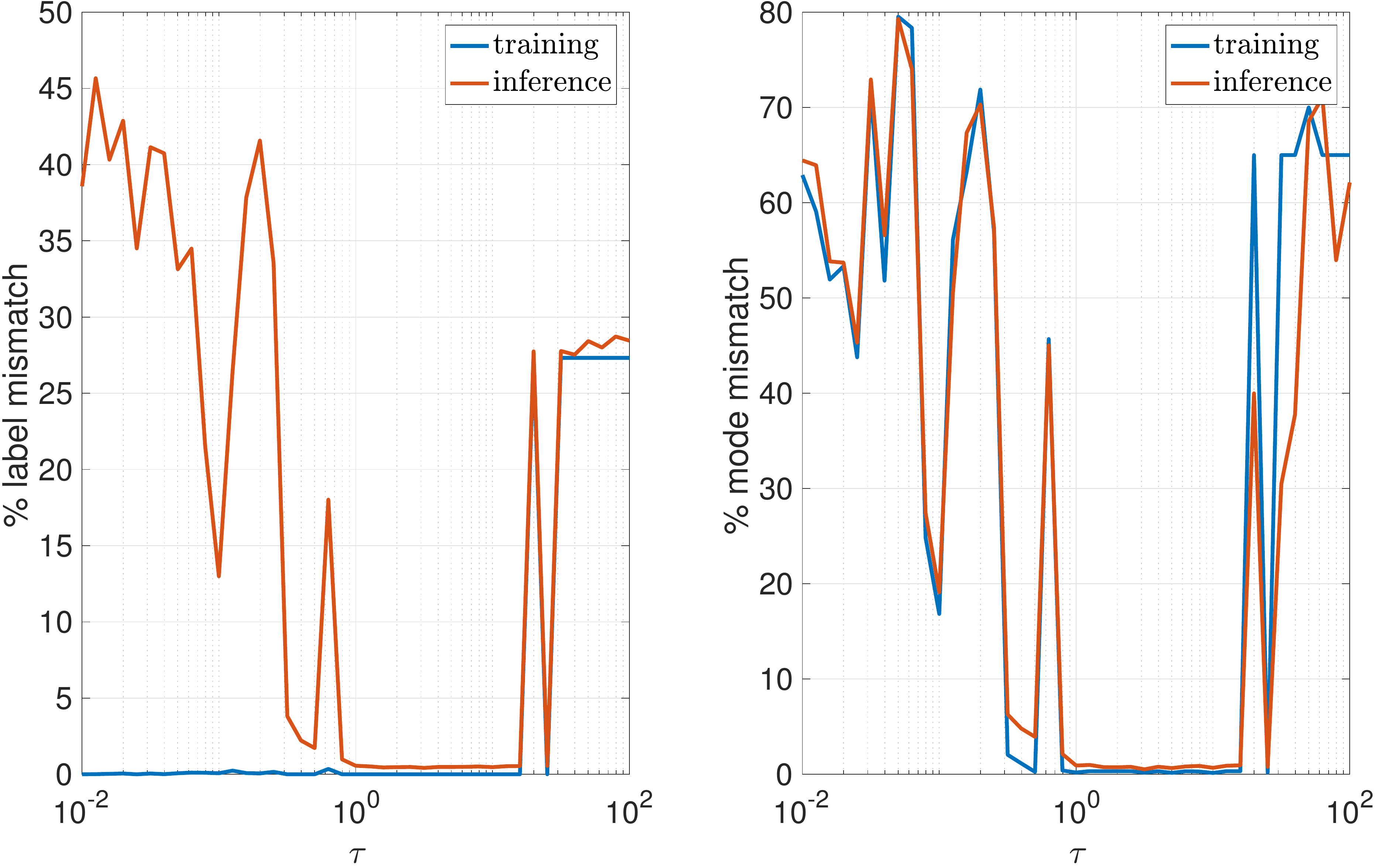}
\caption{Jump binary classifier: percentage of misclassified
labels (left) and mode mismatch (right) on training and production data}
\label{fig:jm-classifier}
\end{figure}

We want to train a binary classifier defined by the following losses
\beqarno
    \ell(x_t,y_t,\theta_{s_t})&=&\max(1-y_t\theta_{s_t}'x_t,0)\\
r(\theta_k)&=&10^{-5}\|\theta_k\|_2^2\\
\LLtrans(s_t,s_{t-1})&=&\tau(1-\delta_{i,j})\\
\LLinit(s_0)&=&0\\
\LLmode(s_t)&=&0.
\eeqarno%
Figure~\ref{fig:jm-classifier} shows the results obtained for different
values of the hyper-parameter $\tau$. We consider the mismatch between
the true labels $y_t$ and the estimated labels
$y^\star_t=\sign((\theta^\star_{s^\star_t})'x_t)$ returned by 
Algorithm~\ref{algo:JM-fit} on the training data,
and also between the true labels $\tilde y_t$ and the labels
$\hat y_t=\sign((\theta^\star_{\hat s_t})'\tilde x_t)$ returned by
Algorithm~\ref{algo:JM-inference_rec} on the production data.
In addition, we consider the detection of model changes, comparing 
the true modes $s_t$, $\tilde s_t$
and their corresponding estimates $s^\star_t$, $\hat s_t$. 
Good values for $\tau$ are in the range $1\div 10$,
for which model changes are correctly detected
on both training and production data. 

The CPU time for executing Algorithm~\ref{algo:JM-fit} ranges between 
$4.24$ and $80.76$~s, 
with Step~\ref{algo1:model_fit} computed using the QP solver of 
GUROBI 7.02~\cite{gurobi}. Algorithm~\ref{algo:JM-fit} requires between 15 and 199 iterations.   Algorithm~\ref{algo:JM-inference_rec} takes an average
of $0.57~\mu$s per data point for inference.

\subsection{Markov jump linear dynamical system}
We consider the Markov jump linear dynamical system with $K=4$ modes
\[
    x_{t+1}=A_{s_t}x_t+B_{s_t}u_t+\zeta_t
\]
where $x_t,\zeta_t\in\rr^8$, $u_t\in\rr^2$ takes random values in $\{-1,1\}^2$,  $\zeta^j_t\sim\NN(0,\sigma_y)$
for all $j=1,\ldots,8$, the matrix pairs $(A_i,B_i)$ are random stable
systems for all $i=1,\ldots,K$. The modes $s_t$ 
are randomly generated according to an (unknown) transition probability
matrix $\Pi\in\rr^{4\times 4}$. The goal is to estimate the system 
matrices $(A_i,B_i)$, $i=1,\ldots,K$, and the transition probability $\Pi$
from $T=50000$ data pairs $(x_t,u_t)$ available for training,
and validate the results on $\tilde T=50000$ new samples. 
 
Algorithm~\ref{algo:JM-fit} is executed $N=5$ times on the training data
with loss function $\|x_{t+1}-A_{s_t}x_t-B_{s_t}u_t\|_2^2$,
uniform mode transition loss $\LLtrans(i,j)=\tau$,  zero
losses $\LLinit$, $\LLmode$, and regularization $r(\theta_k)=10^{-5}\|\theta_k\|_2^2$.
Note that, since the output sample $y_t=x_{t+1}$ is multidimensional, we cannot train
a model for each component of $y$ independently, as they are linked 
by the common mode $s_t$. 

After training and before performing inference via~\eqref{eq:incremental-predictor-eq},
the transition probability matrix $\hat\Pi$ is reconstructed using~\eqref{eq:a-posteriori-mode-loss} on the estimated mode sequence $S^\star$ returned 
by Algorithm~\ref{algo:JM-fit}.

\begin{figure}
\includegraphics[width=\hsize]{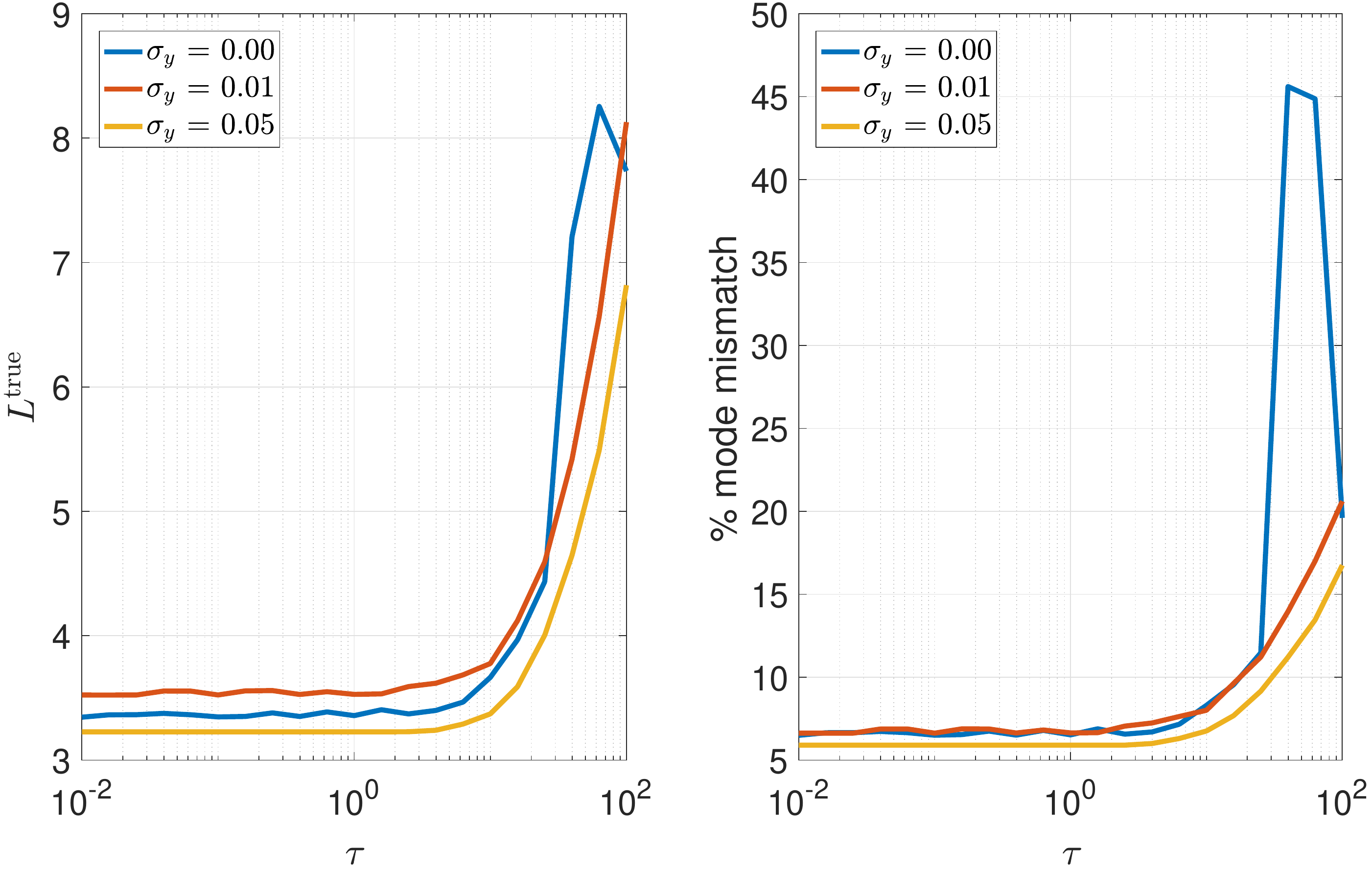}
\caption{Markov jump linear dynamical system: true loss $\tilde L^{\rm true}$ (left)
and mode mismatch $\ell_s^{\rm true}$ (right).} \label{fig:jm-MJLS}
\end{figure}

The results are reported in Figure~\ref{fig:jm-MJLS}. The coefficients of the 
models $(A_i,B_i)$ are estimated with an error of $10^{-8}$ ($\sigma_y=0$),
$10^{-3}$ ($\sigma_y=0.01$), and $10^{-2}$ ($\sigma_y=0.05$), respectively, 
while the transition probability matrix with error $\|\Pi-\hat\Pi\|_2$ of $0.01$ 
for all values of $\sigma_y$. The average
CPU time for executing Algorithm~\ref{algo:JM-fit} is $68$~ms (the longest execution
takes 134 iterations), while Algorithm~\ref{algo:JM-inference_rec} takes
 $0.57~\mu$s per data point on average for inference.

\subsection{Experimental example: PWA dynamical model}
We consider the problem of modeling the dynamics of a placement process of electronic components in a pick-and-place machine described in~\cite{JHF04}. 
The process  consists  of  a  mounting  head  carrying  the  electronic
component which is placed on a printed circuit board, and then released.  This process is characterized by two main operating modes, the \emph{free} and the \emph{impact mode}.
In free mode the  machine carries the electronic component in an unconstrained environment, i.e., without being in contact with the circuit board. In impact mode the mounting head moves in contact with the circuit board.  Because of its switching behaviour, this process has been used as a benchmark to assess the performance of several identification algorithms for hybrid dynamical systems~\cite{BGPV05,JHFVPN05,OhLj13}.

A data record over an interval of $15$~s is gathered from an experimental bench (see~\cite{JHF04} for details), with a sampling frequency of $400$~Hz. We denote by $u$ the voltage applied to the motor driving the mounting head and by $y$ the vertical position of the mounting head. The data record is split in two disjoint subsets: a training set with $T=4800$~samples, which consist of the observations gathered in the first $12$~s of the experiments, and a test set with $\tilde{T}=1200$~samples, which consist of the observations gathered in the last $3$~s.

We want to fit a PWA model as defined in~\eqref{eq:pwa}--\eqref{eq:pwainference} with $K=2$ discrete modes. Each regression model is given by $y_t = \theta_{y,s_t}' \smallmat{x_t\\1}$, where $x_t = \left[y_{t-1} \ \ y_{t-2} \ \ u_{t-1} \ \ u_{t-2} \right]'$.

Algorithm~\ref{algo:JM-fit} is executed $N=5$ times on the first $4400$ samples of the training set with loss function $\ell(x_t,y_t,\theta_{s_t})$ as in~\eqref{eq:pwa},  mode sequence loss $\LL=0$  and regularization $r(\Theta)=\sum_{k=1}^Kr(\theta_{y,k})$, with $r(\theta_{y,k})=10^{-5}\|\theta_{y,k}\|_2^2$. The remaining $400$ samples are used to tune the hyper-parameter $\hyperPWA$ in~\eqref{eq:pwa}, leading to an optimal value $\hyperPWA=2.15\cdot10^{-4}$. The average CPU time for executing  Algorithm~\ref{algo:JM-fit} for a fixed value of $\hyperPWA$ is $156$~ms. In the worst case, Algorithm~\ref{algo:JM-fit} terminates after $25$ iterations.

\begin{figure}
	\includegraphics[width=\hsize]{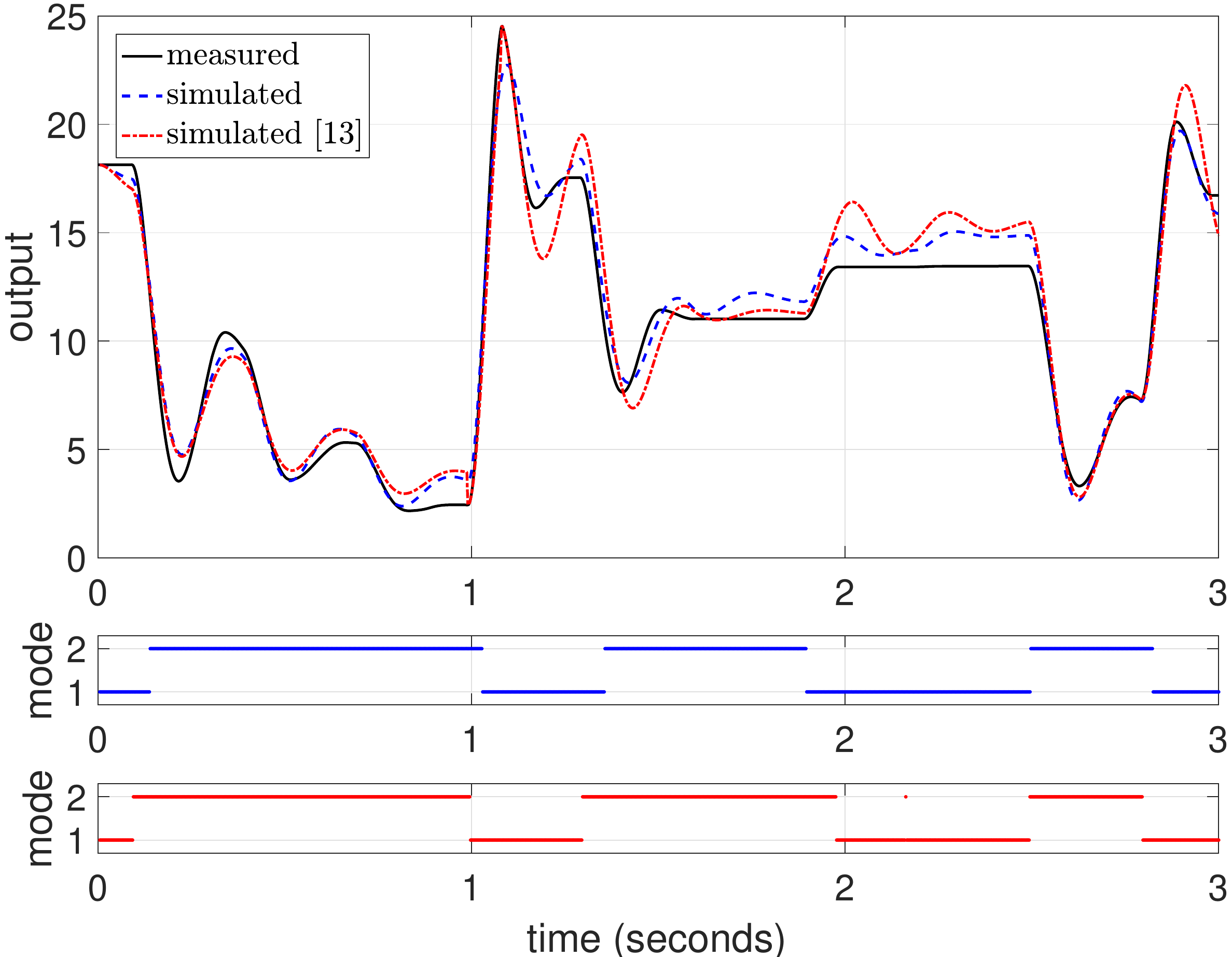}
	\caption{Pick-and-place machine: simulated and actual output (top), mode sequence estimated using our approach (middle), and using the cluster-based algorithm~\cite{FeMuLiMo03} (bottom).} \label{fig:jm-PPM}
\end{figure}

Figure~\ref{fig:jm-PPM} shows the outputs $\tilde y_t$ collected from the production dataset,
the open-loop prediction $\hat y_t$ of the output reconstructed by feeding the same inputs $\tilde u_t$ to the estimated PWA model, and the
sequence of estimated modes $\hat s_t$. The resulting
\emph{best fit rate} $\textrm{BFR} = 100 \left(1- \sqrt{ \frac{  
\sum_{t=1}^{\tilde{T}}  \|\tilde y_t - \hat{y}_t\|^2	
 }{\sum_{t=1}^{\tilde{T}}  \|\tilde y_t - \bar{y}\|^2 }  } \right) \%
$ is equal to $83$\%, where $\bar{y}$ denotes the average of the outputs $\tilde y_1,\ldots,\tilde y_{\tilde T}$. The evolution of the reconstructed mode sequence shows that mode $1$ is active at, roughly,  $y\geq 15$. From the physical knowledge of the system and of the experimental setup, we can associate mode $1$ and $2$ to the  impact and to the free mode, respectively.
 
For comparison, the same fitting problem is solved by using the cluster-based  algorithm for PWA regression in~\cite{FeMuLiMo03}, using the \emph{Hybrid Identification Toolbox} (HIT) toolbox~\cite{HIT}. The \emph{Proximal Support Vector Classifier} (PSVC)~\cite{FM05} 
is employed to compute the polyhedral partition of the regressor space. The same training and production datasets are considered, with the hyper-parameters characterizing the PWA regression algorithm~\cite{FeMuLiMo03} tuned via cross-validation on the last 400 samples of the training set. The open-loop predicted output $\hat y_t$ is shown in Figure~\ref{fig:jm-PPM}, along with the estimated mode sequence. The achieved \textrm{BFR} is 75\%, which is slightly worse
than what we obtained using our approach (83 \%), although very similar.
The average  CPU time required by the HIT toolbox to train the PWA  model for fixed hyper-parameters is  $159$~s, with is about $1000$x
slower than the method proposed in this paper.

\section{Conclusions}
We have presented a new framework for fitting a jump model to a temporal sequence of data. Overall, the approach is able to fit models with latent discrete variables and provides 
an efficient (and more general) alternative to existing methods, such as the expectation-maximization algorithm for the estimation of hidden Markov models and cluster-based heuristics for the identification of switching and PWA models.
 
A main strength of the approach is its versatility in describing a large class of parametric models, as the shape of the model and the way it jumps depends on the shape of the loss functions used for fitting the model parameters and for inference. Such a generality of the approach
stimulates future research to address auto-tuning strategies, where the loss functions are chosen automatically from data. We expect that several instances of our approach will be investigated, using different loss functions and in various applications.

Another strength of the proposed approach is its numerical efficiency, due to using a simple coordinate-descent optimization algorithm for fitting model parameters
and a recursive formulation for inferring outputs and latent modes. Although there is no
guarantee of converging to the global optimum, 
numerical evidence shows the effectiveness of the method. 

Future research will also  address an incremental version of the fitting algorithm, so to update models and infer output/mode pairs when data are streaming on-line.

\begin{thebibliography}{10}

\bibitem{BPSW70}
L.E. Baum, T.~Petrie, G.~Soules, and N.~Weiss.
\newblock A maximization technique occurring in the statistical analysis of
  probabilistic functions of {Markov} chains.
\newblock {\em The Annals of Mathematical Statistics}, 41(1):164--171, 1970.

\bibitem{Bel57}
R.~Bellman.
\newblock {\em Dynamic Programming}.
\newblock Princeton University Press, Princeton, NJ, USA, 1957.

\bibitem{BGPV05}
A.~Bemporad, A.~Garulli, S.~Paoletti, and A.~Vicino.
\newblock A bounded-error approach to piecewise affine system identification.
\newblock {\em IEEE Trans. Autom. Control}, 50(10):1567--1580, October 2005.

\bibitem{BG06}
A.~Bemporad and N.~Giorgetti.
\newblock Logic-based methods for optimal control of hybrid systems.
\newblock {\em IEEE Transaction on Automatic Control}, 51(6):963--976, 2006.

\bibitem{BRL01a}
A.~Bemporad, J.~Roll, and L.~Ljung.
\newblock Identification of hybrid systems via mixed-integer programming.
\newblock In {\em Proc. 40th IEEE Conf. on Decision and Control}, pages
  786--792, Orlando, Florida, 2001.

\bibitem{BM94}
K.P. Bennett and O.L. Mangasarian.
\newblock Multicategory discrimination via linear programming.
\newblock {\em Optimization Methods and Software}, 3:27--39, 1994.

\bibitem{BV04}
S.~Boyd and L.~Vandenberghe.
\newblock {\em Convex Optimization}.
\newblock Cambridge University Press, New York, NY, USA, 2004.
\newblock \url{http://www.stanford.edu/~boyd/cvxbook.html}.

\bibitem{BrBePi16ECC}
V.~Breschi, A.~Bemporad, and D.~Piga.
\newblock Identification of hybrid and linear parameter varying models via
  recursive piecewise affine regression and discrimination.
\newblock In {\em European Control Conference}, pages 2632--2637, Aalborg,
  Denmark, 2016.

\bibitem{BPB16a}
V.~Breschi, D.~Piga, and A.~Bemporad.
\newblock Piecewise affine regression via recursive multiple least squares and
  multicategory discrimination.
\newblock {\em Automatica}, 73:155--162, November 2016.

\bibitem{chan2008modeling}
A.B. Chan and N.~Vasconcelos.
\newblock Modeling, clustering, and segmenting video with mixtures of dynamic
  textures.
\newblock {\em IEEE transactions on pattern analysis and machine intelligence},
  30(5):909--926, 2008.

\bibitem{costa2006discrete}
O.~L.~V. Costa, M.~D. Fragoso, and R.~P. Marques.
\newblock {\em Discrete-time Markov jump linear systems}.
\newblock Springer Science \& Business Media, 2006.

\bibitem{DLR77}
A.P. Dempster, N.M. Laird, and D.B. Rubin.
\newblock Maximum likelihood from incomplete data via the {EM} algorithm.
\newblock {\em Journal of the Royal Statistical Society, Series B.},
  39(1):1--38, 1977.

\bibitem{HIT}
G.~Ferrari-Trecate.
\newblock Hybrid identification toolbox {(HIT)}, 2005.

\bibitem{FeMuLiMo03}
G.~Ferrari-Trecate, M.~Muselli, D.~Liberati, and M.~Morari.
\newblock A clustering technique for the identification of piecewise affine
  systems.
\newblock {\em Automatica}, 39(2):205--217, 2003.

\bibitem{Fri94}
M.~Fridman.
\newblock Hidden {Markov} model regression.
\newblock Technical report, Institute of Mathematics, University of Minnesota,
  Minneapolis, MN, 1994.

\bibitem{FM05}
G.M. Fung and O.L. Mangasarian.
\newblock Multicategory proximal support vector machine classifiers.
\newblock {\em Machine Learning}, 59:77--97, 2005.

\bibitem{Gu2011markov}
M.~Guidolin.
\newblock Markov switching models in empirical finance.
\newblock In {\em Missing Data Methods: Time-Series Methods and Applications},
  pages 1--86. Emerald Group Publishing Limited, 2011.

\bibitem{gurobi}
{Gurobi Optimization, Inc.}
\newblock {\em Gurobi Optimizer Reference Manual}, 2017.

\bibitem{HTF09}
T.~Hastie, R.~Tibshirani, and J.~Friedman.
\newblock {\em The elements of statistical learning}.
\newblock Springer, New York, 2nd edition, 2009.

\bibitem{JWH04}
A.~Juloski, S.~Weiland, and M.Heemels.
\newblock A {Bayesian} approach to identification of hybrid systems.
\newblock In {\em Proc. 43th IEEE Conf. on Decision and Control}, Paradise
  Island, Bahamas, 2004.

\bibitem{JHF04}
A.L. Juloski, W.P.M.H. Heemels, and G.~Ferrari-Trecate.
\newblock Data-based hybrid modelling of the component placement process in
  pick-and-place machines.
\newblock {\em Control Engineering Practice}, 12(10):1241--1252, 2004.

\bibitem{JHFVPN05}
A.L. Juloski, W.P.M.H. Heemels, G.~Ferrari-Trecate, R.Vidal, S.~Paoletti, and
  J.H.G. Niessen.
\newblock Comparison of four procedures for the identification of hybrid
  systems.
\newblock {\em Lecture Notes in Computer Science}, 3414:354--369, 2005.

\bibitem{manning2008introduction}
C.D. Manning, P.~Raghavan, and H.~Sch{\"u}tze.
\newblock {\em Introduction to information retrieval}, volume~1.
\newblock 2008.

\bibitem{OhSa2008}
S.M. Oh, J.M Rehg, T.~Balch, and F.~Dellaert.
\newblock Learning and inferring motion patterns using parametric segmental
  switching linear dynamic systems.
\newblock {\em International Journal of Computer Vision}, 77(1):103--124, 2008.

\bibitem{OhLj13}
H.~Ohlsson and L.~Ljung.
\newblock Identification of switched linear regression models using
  sum-of-norms regularization.
\newblock {\em Automatica}, 49(4):1045--1050, 2013.

\bibitem{OzSzLa2010}
N.~Ozay, Mario M.~Sznaier, and C.~Lagoa.
\newblock Model (in)validation of switched arx systems with unknown switches
  and its application to activity monitoring.
\newblock In {\em 49th IEEE Conference on Decision and Control}, pages
  7624--7630, Atlanta, GA, 2010.

\bibitem{PaReMac2001}
V.~Pavlovic, J.M. Rehg, and J.~MacCormick.
\newblock Learning switching linear models of human motion.
\newblock In {\em Advances in neural information processing systems}, pages
  981--987, 2001.

\bibitem{Pil16}
G.~Pillonetto.
\newblock A new kernel-based approach to hybrid system identification.
\newblock {\em Automatica}, 70:21--31, 2016.

\bibitem{Rab89}
L.R. Rabiner.
\newblock A tutorial on hidden {Markov} models and selected applications in
  speech recognition.
\newblock {\em Proceedings of the IEEE}, 77(2):257--286, 1989.

\bibitem{ShWo2008}
B.~Schuller, M.~W{\"o}llmer, T.~Moosmayr, G.~Ruske, and G.~Rigoll.
\newblock Switching linear dynamic models for noise robust in-car speech
  recognition.
\newblock {\em Pattern Recognition}, pages 244--253, 2008.

\bibitem{TiMar15}
A.~Timmermann.
\newblock {\em Markov Switching Models in Finance}, volume~4.
\newblock John Wiley \& Sons, Ltd, 2015.

\bibitem{VCS02}
R.~Vidal, A.~Chiuso, and S.~Soatto.
\newblock Observability and identifiability of jump linear systems.
\newblock In {\em Proc. 41st IEEE Conference on Decision and Control},
  volume~4, pages 3614--3619, 2002.

\bibitem{viterbi2010error}
A.~J. Viterbi.
\newblock Error bounds for convolutional codes and an asymptotically optimum
  decoding algorithm.
\newblock In {\em The Foundations Of The Digital Wireless World: Selected Works
  of AJ Viterbi}, pages 41--50. 2010.

\end{thebibliography}
\end{document}